\documentclass{article} % For LaTeX2e
\usepackage{iclr2027_conference}
\usepackage{times}
\usepackage{float}
\usepackage{placeins}
\usepackage{amsmath,amsfonts,bm}

\def\eqref#1{equation~\ref{#1}}
\def\1{\bm{1}}

\DeclareMathAlphabet{\mathsfit}{\encodingdefault}{\sfdefault}{m}{sl}
\SetMathAlphabet{\mathsfit}{bold}{\encodingdefault}{\sfdefault}{bx}{n}

\newcommand{\R}{\mathbb{R}}

\DeclareMathOperator*{\argmin}{arg\,min}

\usepackage{amsmath,amssymb,amsthm,mathtools}
\usepackage{bm}
\usepackage{xcolor}
\usepackage{url}
\usepackage{caption}
\usepackage{makecell}
\usepackage{graphicx}
\usepackage{booktabs}
\usepackage{algorithm}
\usepackage{algpseudocode}
\usepackage{tabularx}
\usepackage{multirow, multicol}

\usepackage{tikz}
\usepackage{pgfplots}
\usepgfplotslibrary{groupplots}
\pgfplotsset{compat=1.18}
\usetikzlibrary{patterns, shapes.geometric, calc, arrows.meta}
\usepackage{wrapfig}
\usepackage{rotating}
\usepackage{calc}
\usepackage{xspace}
\usepackage{hyperref}

\hypersetup{
    colorlinks=true,
    citecolor=blue,
    linkcolor=blue,
    urlcolor=blue
}

\newcommand{\appref}[1]{\hyperref[#1]{Appendix~\ref*{#1}}}

\title{Prune Gradually, Adapt Sparsely:\\Efficient Pruning of Large Language Models at Extreme Sparsity}
\newcommand{\etapsauthorcell}[3]{%
  \begin{minipage}[t]{#1\linewidth}
    \centering\normalfont
    \parbox[b][2.6em][b]{\linewidth}{%
      \centering\bfseries #2%
    }\par
    \vspace{0.2em}
    #3
  \end{minipage}%
}

\author{%
\begin{minipage}[t]{\dimexpr\textwidth-2\tabcolsep\relax}
\centering\normalfont
\etapsauthorcell{0.36}
  {Jean Lucien Randrianantenaina\thanks{Correspondance to: \href{mailto:jlr85@bath.ac.uk}{jlr85@bath.ac.uk}}}
  {University of Bath}
\hfill
\etapsauthorcell{0.31}
  {Hung Nguyen}
  {Aalto University}
\hfill
\etapsauthorcell{0.31}
  {Nasib Ullah}
  {Aalto University}
\hfill
\\
\etapsauthorcell{0.22}
  {Jinbin Zhang}
  {Aalto University}
\etapsauthorcell{0.31}
  {Nilesh Gupta}
  {University of Texas at Austin}
\hfill
\etapsauthorcell{0.22}
  {Inderjit S. Dhillon}
  {Google}
\hfill
\etapsauthorcell{0.22}
  {Rohit Babbar}
  {University of Bath}
\\[0.8em]
\end{minipage}%
}

\newcommand{\Smax}{S_{\max}}
\newcommand{\TopK}{\operatorname{TopK}}
\newcommand{\supp}{\operatorname{supp}}
\newcommand{\loss}{\mathcal{L}}
\newcommand{\dth}{d_\theta}
\newcommand{\dph}{d_\phi}
\newcommand{\dphi}[1]{d_{\phi,#1}}
\newcommand{\dense}{\textsc{Dense}\xspace}
\newcommand{\etaps}{\textsc{Etaps}\xspace}
\newcommand{\elsa}{\textsc{Elsa}\xspace}
\newcommand{\seft}{\textsc{Seft}\xspace}
\newcommand{\lora}{\textsc{LoRA}\xspace}
\newcommand{\alps}{\textsc{ALPS}\xspace}
\newcommand{\magnitude}{\textsc{Magnitude}\xspace}
\newcommand{\sparsegpt}{\textsc{SparseGPT}\xspace}
\newcommand{\wanda}{\textsc{Wanda}\xspace}
\newcommand{\pc}[1]{#1\%\xspace}

\iclrfinalcopy 
\begin{document}

\maketitle
\pagestyle{fancy}
\fancyhead[L]{Preprint}
\begin{abstract}
Weight pruning can reduce the storage and inference cost of large language models (LLMs), but preserving quality at high sparsity levels remains challenging. Existing methods face a quality--cost trade-off in this regime. One-shot pruning is cheap but degrades sharply as sparsity increases, even when followed by post-pruning recovery. A growing line of work instead optimizes the mask directly from the dense model under the end-to-end objective; among these methods, \elsa~\citep{lee2026elsa}, the recent state of the art at extreme sparsity, reaches high quality but at the cost of model-sized optimizer state and auxiliary variables. To address this gap, we propose \etaps, which prunes gradually and adapts sparsely: it tightens a global sparsity constraint in progressive stages while training a sparse delta with a small, fixed budget. Our key observation is that gradual weight removal keeps each pruning step small enough for sparse adaptation to compensate, allowing the adapted model to provide end-to-end gradients that guide the next pruning decision without model-sized optimizer state. Across six models from 125M to 13B parameters, \etaps achieves lower perplexity than \elsa at every evaluated sparsity from \pc{50} to \pc{90}, with perplexity reductions of up to \pc{42}. \etaps is also far cheaper: at \pc{90} sparsity, it prunes LLaMA-2-7B on a single H200, where \elsa uses four, with $5.0\times$ smaller total persistent state and \pc{49} fewer GPU-hours. Together, our findings point to a promising direction for achieving high quality in demanding regimes such as extreme sparsity at substantially lower cost.
\end{abstract}

\section{Introduction}
\label{sec:introduction}

Large language models (LLMs) are expensive to deploy because they require moving billions of parameters through the memory hierarchy during inference. This is especially true during autoregressive decoding, where each token requires another pass over the weights, making execution memory-bandwidth bound at typical deployment batch sizes \citep{xia2023flashllm, macko2026spmv}. Weight sparsity offers a direct remedy, as it reduces parameter storage and weight traffic. When paired with a suitable representation and kernel, these reductions can translate into lower inference cost \citep{xia2023flashllm, okanovic2025blast}: with the MACKO SpMV kernel \citep{macko2026spmv}, a \pc{90}-sparse LLaMA-2-7B runs with $2.5\times$ lower latency and $4.6\times$ less memory on an RTX~3090 \citep{lee2026elsa}. Since the savings grow with the fraction of weights removed, higher sparsity is particularly desirable. The difficulty is that pruning becomes least forgiving exactly where it is most rewarding, as conventional methods degrade sharply at high compression (\pc{90}--\pc{99}) \citep{frantar2023sparsegpt, sun2024wanda, lee2026elsa}. Our work targets this extreme sparsity regime, where recovering model quality is most challenging and existing methods are pushed to their limits.

One-shot pruning provides the most economical route into sparsity. Methods such as Wanda~\citep{sun2024wanda}, SparseGPT~\citep{frantar2023sparsegpt}, and Pruner-Zero~\citep{dong2024prunerzero} construct sparse models directly from pretrained checkpoints using a small calibration set, with pruning decisions made from a cheap criterion without retraining. Although one-shot pruning remains robust at moderate compression (\pc{{50}}--\pc{{60}}), its performance degrades significantly at higher sparsity levels \citep{lee2026elsa}. One common explanation is that most one-shot pruning criteria favor locally optimal solutions that are misaligned with the end-to-end objective, causing errors from individual pruning decisions to accumulate as sparsity increases \citep{lee2026elsa, mozaffari2026leap}.

A natural fix is to recover the model after pruning, allowing the surviving parameters to compensate for pruning-induced errors via end-to-end loss gradients \citep{hu2022lora,lu2024spp, munoz2024sqft, zimmer2023perp, xiao2025seft}. While most recovery methods update only the remaining weights, \seft~\citep{xiao2025seft} further allows the sparse topology to evolve, steering the model toward sparse configurations better aligned with the fine-tuning objective. However, at extreme sparsity levels (\pc{90}--\pc{99}), we observe that recovered models degrade sharply as sparsity grows (\autoref{tab:extreme}). Recovery also remains sensitive to its starting point: even \seft, despite being able to adjust its topology, reaches noticeably different quality depending on the pruning method used to initialize it. This indicates that although post-pruning recovery can improve a sparse model, its outcome is heavily constrained by the quality of the initial sparse state.

A complementary line of work constructs the mask directly from the dense model, using the language-modeling objective to capture the global effect of weight removal \citep{lee2026elsa, fang2024maskllm, mozaffari2026leap, hourri2025patch}. Among these, \elsa~\citep{lee2026elsa}, our primary comparator, achieves strong quality in the extreme-sparsity regime by jointly optimizing the weights and mask under the end-to-end loss using ADMM. However, it requires model-sized auxiliary variables on top of optimizer state, exceeding the persistent memory of dense fine-tuning at matched precision (\autoref{tab:cost}). As a result, \elsa illustrates the other end of the quality--cost trade-off, where strong quality at extreme sparsity comes at the expense of substantially higher memory cost.

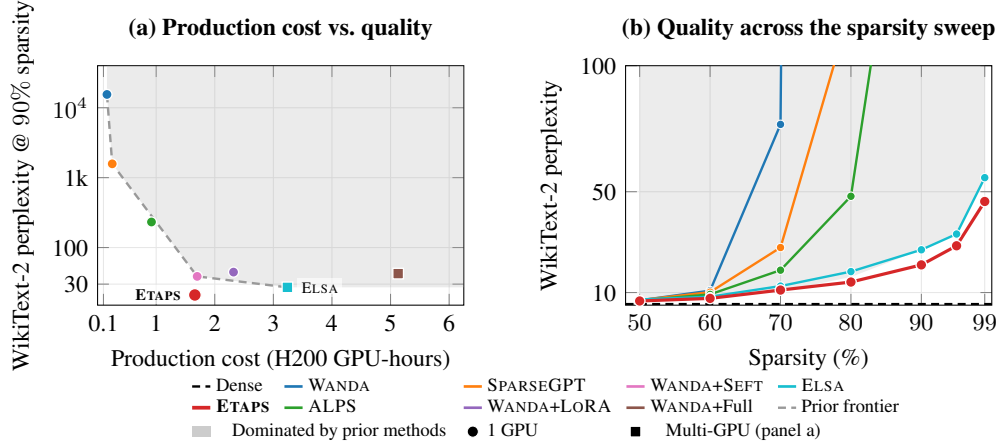
\begin{figure}[t]
\centering
\begin{tikzpicture}
\definecolor{cWanda}{HTML}{1f77b4}
\definecolor{cSparseGPT}{HTML}{ff7f0e}
\definecolor{cALPS}{HTML}{2ca02c}
\definecolor{cLoRA}{HTML}{9467bd}
\definecolor{cFull}{HTML}{8c564b}
\definecolor{cSEFT}{HTML}{e377c2}
\definecolor{cELSA}{HTML}{17becf}
\definecolor{cETAPS}{HTML}{d62728}
\definecolor{cGuide}{HTML}{999999}
\tikzset{lbl/.style={font=\tiny, inner sep=1.5pt, fill=white, fill opacity=0.6, text opacity=1}}

% ---------------------------------------------------------------- panel (a)
% Values: GPU-hours from Table 5 (tab:cost), WikiText-2 at 90% from Table 3 (tab:extreme).
% Circles = single-GPU run, squares = multi-GPU run (Table 5).
% TODO: replace the A100 timings of Wanda (0.16), SparseGPT (0.25),
% taken from Lee et al. (2026), with our H200 measurements.
\begin{axis}[
    name=axA, width=0.46\textwidth, height=0.34\textwidth,
     ymode=log,
    xlabel={Production cost (H200 GPU-hours)},
    ylabel={WikiText-2 perplexity @ $90\%$ sparsity},
    xlabel style={font=\small}, ylabel style={font=\small},
    tick label style={font=\footnotesize},
    xmin=0, xmax=6.25, ymin=15, ymax=40000,
    xtick={0.1,1,2,3,4,5,6}, xticklabels={0.1,1,2,3,4,5, 6},
    ytick={30,100,1000,10000}, yticklabels={30,100,1k,$10^4$},
    grid=both, grid style={line width=0.25pt,draw=gray!20},
    title={(a) Production cost vs.\ quality},
    title style={font=\small\bfseries},
    clip=true,
]
% Prior-method Pareto frontier (ETAPS excluded from its construction); the
% shaded region holds outcomes dominated by that frontier.
\addplot[fill=gray!65, fill opacity=0.24, draw=none, forget plot] coordinates {
    (0.16,15507) (0.25,1577) (1.70,38.71) (3.24,26.97)
    (20,26.97) (20,40000) (0.16,40000)} -- cycle;
\addplot[color=cGuide, densely dashed, line width=0.9pt, mark=none, forget plot]
    coordinates {(0.16,15507) (0.25,1577) (1.70,38.71) (3.24,26.97)};

% single-GPU runs
\addplot[only marks, mark=*, mark size=1.9pt,
    mark options={fill=cWanda,draw=white,line width=0.5pt}, forget plot]
    coordinates {(0.16,15507)};
\addplot[only marks, mark=*, mark size=1.9pt,
    mark options={fill=cSparseGPT,draw=white,line width=0.5pt}, forget plot]
    coordinates {(0.25,1577)};
\addplot[only marks, mark=*, mark size=1.9pt,
    mark options={fill=cALPS,draw=white,line width=0.5pt}, forget plot]
    coordinates {(0.92,232.20)};
\addplot[only marks, mark=*, mark size=1.9pt,
    mark options={fill=cLoRA,draw=white,line width=0.5pt}, forget plot]
    coordinates {(2.32,44.57)};
\addplot[only marks, mark=*, mark size=1.9pt,
    mark options={fill=cSEFT,draw=white,line width=0.5pt}, forget plot]
    coordinates {(1.70,38.71)};
% multi-GPU runs: Wanda+Full (2 GPUs), ELSA (4 GPUs)
\addplot[only marks, mark=square*, mark size=2.0pt,
    mark options={fill=cFull,draw=white,line width=0.5pt}, forget plot]
    coordinates {(5.13,42.40)};
\addplot[only marks, mark=square*, mark size=2.0pt,
    mark options={fill=cELSA,draw=white,line width=0.5pt}, forget plot]
    coordinates {(3.24,26.97)};
% ETAPS (single GPU)
\addplot[only marks, mark=*, mark size=2.4pt,
    mark options={fill=cETAPS,draw=white,line width=0.5pt}, forget plot]
    coordinates {(1.66,20.98)};

\node[lbl, anchor=east] at (axis cs:1.5,20.98) {\textbf{\etaps}};
\node[lbl, anchor=west] at (axis cs:3.4,26.97) {\elsa};
\end{axis}

% ---------------------------------------------------------------- panel (b)
% Values: Table 3 (tab:extreme) and Table 9 (tab:main-ppl-results), LLaMA-2-7B.
\begin{axis}[
    name=axB, width=0.46\textwidth, height=0.34\textwidth, xshift=0.50\textwidth,
    % ymode=log,
    xlabel={Sparsity (\%)},
    ylabel={WikiText-2 perplexity},
    xlabel style={font=\small}, ylabel style={font=\small},
    tick label style={font=\footnotesize},
    xmin=48, xmax=100, ymin=5, ymax=100,
    restrict y to domain=0:30100,
     unbounded coords=jump,
    xtick={50,60,70,80,90,99},
    ytick={10,50,100,300,600}, yticklabels={10,50,100,300,600},
    grid=both, grid style={line width=0.25pt,draw=gray!20},
    title={(b) Quality across the sparsity sweep},
    title style={font=\small\bfseries},
    clip=true
]
% Region above the best prior method at each sparsity.
\addplot[fill=gray, fill opacity=0.15, draw=none, forget plot] coordinates {
    (50,6.85) (60,8.05) (70,10.45) (80,16.16)
    (90,26.97) (95,33.22) (99,55.94) (99,650) (50,650)} -- cycle;

\addplot[color=black, densely dashed, line width=0.8pt, mark=none, forget plot]
    coordinates {(48,5.47) (100,5.47)};

\addplot[color=cWanda, line width=0.9pt, mark=*, mark size=1.6pt,
    mark options={fill=cWanda,draw=white,line width=0.4pt}, forget plot]
    coordinates {(50,6.91) (60,10.76) (70,76.71) (80,3122.23) (90,15507) (95,14757) (99,301000)};

\addplot[color=cSparseGPT, line width=0.9pt, mark=*, mark size=1.6pt,
    mark options={fill=cSparseGPT,draw=white,line width=0.4pt}, forget plot]
    coordinates {(50,6.99) (60,10.16) (70,27.86) (80,123.28) (90,1577) (95,4369)};

% \addplot[color=cSEFT, line width=0.9pt, mark=*, mark size=1.6pt,
%     mark options={fill=cSEFT,draw=white,line width=0.4pt}, forget plot]
%     coordinates {(50, 7.12) (60, 8.05) (70, 10.45) (80, 16.16) (90, 32.89) (95, 56.57) (99, 204.08)}; % best seft

\addplot[color=cALPS, line width=0.9pt, mark=*, mark size=1.6pt,
    mark options={fill=cALPS,draw=white,line width=0.4pt}, forget plot]
    coordinates {(50,6.85) (60,9.28) (70,18.88) (80,48.20) (90,232.20) (95,1093.25) (99,5148.07)};

\addplot[color=cELSA, line width=0.9pt, mark=*, mark size=1.6pt,
    mark options={fill=cELSA,draw=white,line width=0.4pt}, forget plot]
    coordinates {(50,7.27) (60,8.36) (70,12.57) (80,18.32) (90,26.97) (95,33.22) (99,55.58)};

\addplot[color=cETAPS, line width=1.2pt, mark=*, mark size=2.0pt,
    mark options={fill=cETAPS,draw=white,line width=0.5pt}, forget plot]
    coordinates {(50,6.66) (60,7.69) (70,10.99) (80,14.17) (90,20.98) (95,28.56) (99,46.13)};
\end{axis}

% ---------------------------------------------------------------- shared legend
% Method identity by colour; in panel (a) marker shape encodes GPU count.
\node[anchor=north, font=\scriptsize, inner sep=1.5pt]
    at ($(axA.south)!0.5!(axB.south)-(0,0.88cm)$) {
    \begin{tabular}{@{}c@{\hspace{1.5pt}}l@{\hspace{5pt}}c@{\hspace{1.5pt}}l@{\hspace{5pt}}c@{\hspace{1.5pt}}l@{\hspace{5pt}}c@{\hspace{1.5pt}}l@{\hspace{5pt}}c@{\hspace{1.5pt}}l@{}}
      \tikz[baseline=-0.55ex] \draw[black,densely dashed,line width=0.9pt] (0,0)--(0.24,0); & Dense &
      \tikz[baseline=-0.55ex] \draw[cWanda,line width=1.0pt] (0,0)--(0.24,0); & \wanda &
      \tikz[baseline=-0.55ex] \draw[cSparseGPT,line width=1.0pt] (0,0)--(0.24,0); & \sparsegpt &
      \tikz[baseline=-0.55ex] \draw[cSEFT,line width=1.0pt] (0,0)--(0.24,0); & \wanda{}+\seft &
      \tikz[baseline=-0.55ex] \draw[cELSA,line width=1.0pt] (0,0)--(0.24,0); & \elsa \\
      \tikz[baseline=-0.55ex] \draw[cETAPS,line width=1.4pt] (0,0)--(0.24,0); & \textbf{\etaps} &
      \tikz[baseline=-0.55ex] \draw[cALPS,line width=1.0pt] (0,0)--(0.24,0); & \alps &
      \tikz[baseline=-0.55ex] \draw[cLoRA,line width=1.0pt] (0,0)--(0.24,0); & \wanda{}+\lora &
      \tikz[baseline=-0.55ex] \draw[cFull,line width=1.0pt] (0,0)--(0.24,0); & \wanda{}+Full &
      \tikz[baseline=-0.55ex] \draw[cGuide,densely dashed,line width=0.9pt] (0,0)--(0.24,0); & Prior frontier \\[1pt]
      \tikz[baseline=-0.55ex] \draw[fill=gray!40,draw=none] (0,-0.07)rectangle(0.24,0.07); & \multicolumn{3}{l}{Dominated by prior methods} &
      \tikz[baseline=-0.55ex] \draw[fill=black,draw=white,line width=0.4pt] (0.12,0) circle (2.1pt); & 1 GPU &
      \tikz[baseline=-0.55ex] \draw[fill=black,draw=white,line width=0.4pt] (0.05,-0.07) rectangle (0.19,0.07); & \multicolumn{3}{l}{Multi-GPU (panel a)}
    \end{tabular}
};
\end{tikzpicture}
\caption{Cost--quality trade-off for  sparse LLaMA-2-7B. Both panels report WikiText-2
perplexity.}
\label{fig:cost-quality}
\end{figure}

This leaves an open question: \textit{can extreme-sparsity pruning be guided by gradients from the end-to-end loss throughout its trajectory without paying for model-sized optimizer state?} Existing approaches reveal a trade-off in this regime: inexpensive methods select the sparse support with limited or no gradient information, whereas strong model-level optimization can require dense-model state overhead. \etaps addresses this gap by bringing together two ingredients for extreme-sparsity LLM pruning: (i) a gradual sparsification trajectory \citep{zhu2018prune}, which avoids committing to an extreme mask in a single step, and (ii) a sparse delta \citep{ansell2024spiel, xiao2025seft} that carries small trainable and optimizer state, making end-to-end adaptation affordable throughout pruning. Our key observation is that these two ingredients support each other at extreme sparsity. The gradual trajectory makes a small adaptation budget feasible: a model with few trainable parameters struggles to recover from losing most of its weights at once, but recovers better when they are removed a fraction at a time. Adaptation, in turn, informs pruning: gradient statistics gathered at each stage guide the next pruning step, so each decision is built on a model already adapted to the preceding sparsity level. Our ablations show that both advantages grow with sparsity, as progressive pruning outperforms direct projection and gradient-weighted scoring outperforms magnitude alone. Together, they let \etaps traverse the path to extreme sparsity at a fraction of full-model optimizer cost while achieving strong quality in this regime (\autoref{fig:cost-quality}). 

We empirically demonstrate the scalability and efficiency of \etaps across LLM families and model sizes, from 125M to 70B parameters. Across six models, \etaps achieves lower perplexity than \elsa at every tested sparsity from \pc{50} to \pc{90}. On LLaMA-2-7B, \etaps achieves the lowest perplexity of all evaluated methods across \pc{50}--\pc{99} sparsity, and notably, in the extreme regime (\pc{90}--\pc{99}) it reduces perplexity relative to \elsa by \pc{14.0}--\pc{25.0}. Moreover, as the small delta carries only sparse optimizer state, \etaps also requires far less memory: it prunes LLaMA-2-7B on a single H200 with $5.0\times$ less persistent state and \pc{49} fewer GPU-hours than \elsa on four equivalent GPUs. Finally, \etaps scales to 67B--70B models and extends naturally to $N{:}M$ sparsity.

Our contributions are threefold:
\begin{itemize}
\itemsep2pt

\item \textbf{Progressive pruning with increasing adaptation intervals.}
\etaps reaches the target sparsity gradually, using longer adaptation intervals at later pruning stages. Controlled comparisons with direct projection and uniform step allocation show that both the progressive trajectory and the adaptation schedule affect quality at extreme sparsity.

\item \textbf{Parameter-efficient model-level adaptation.}
\etaps adapts the model using a sparse delta trained on the end-to-end loss. Its persistent trainable parameters and optimizer state scale with a fixed budget, far smaller than the model ($\dph \ll \dth$). It uses gradient statistics to select the delta's coordinates, requiring only one dense, moment-free accumulator.

\item \textbf{Quality, efficiency, and scale.}
\etaps demonstrates that extreme-sparsity pruning can achieve competitive quality and a favorable quality–cost trade-off at LLM scale. It remains strong at \pc{90} sparsity on models up to 70B parameters, extends to \pc{95}--\pc{99} on LLaMA-2-7B, and requires substantially less persistent memory than \elsa.

\end{itemize}

\section{Problem Statement}
\label{sec:problem}

Let $\theta\in\mathbb{R}^{\dth}$ denote the prunable parameters of a pretrained Transformer, and let $\loss_{\mathcal D}(w)$ denote the end-to-end language-modeling objective (next-token cross-entropy) on a dataset $\mathcal D$. For a target $\Smax$, we define the maximum number of surviving parameters $K$ and the corresponding feasible set $\mathcal C_K$ as:
\begin{equation}
    K
    =
    \left\lfloor
        (1-\Smax)\dth
    \right\rfloor,
    \qquad
    \mathcal C_K
    =
    \left\{
        w\in\mathbb{R}^{\dth}
        :
        \|w\|_0\le K
    \right\}.
    \label{eq:target-feasible-set}
\end{equation}
An ideal sparse model satisfies the constrained optimization problem:
\begin{equation}
    w_K^\star
    \in
    \argmin_{w\in\mathcal C_K}
    \loss_{\mathcal D}(w).
    \label{eq:global-sparse-objective}
\end{equation}
Directly optimizing this objective is intractable at LLM scale because support selection is combinatorial and jointly optimizing the surviving weights requires model-scale computation.

Rather than following standard one-shot pruning methods that project the dense model directly into $\mathcal C_K$, we decompose the optimization into a sequence of nested constraints:
\begin{equation}
    \mathcal C_{K_0}
    \supset
    \mathcal C_{K_1}
    \supset
    \cdots
    \supset
    \mathcal C_{K_P},
    \qquad
    \dth
    =K_0>K_1>\cdots>K_P=K.
    \label{eq:constraint-chain}
\end{equation}
Between consecutive constraints, the model is adapted to the current support; this adaptation compensates for the previous pruning step and supplies gradients from the language-modeling objective that determine the next pruning decision. For $N{:}M$, the same principle is applied to group-wise feasible sets. \etaps realizes this by freezing the pretrained weights, adapting a sparse delta with a model-dependent budget $d_\phi$ fixed throughout pruning, and progressively prune using the collected gradient statistics.

\section{Method: \etaps}
\label{sec:method}

\subsection{Overview and Progressive Pruning}
\label{sec:method-overview}

\etaps constructs sparse models progressively over $P$ stages. At stage $i\in\{1,\ldots, P\}$, the effective weights are given by
\begin{equation}
    w_i = M_i\odot(\theta+\delta_i),
\end{equation}
where pretrained weights $\theta\in\R^{\dth}$ remain frozen during all stages. $M_i\in\{0,1\}^{\dth}$ is a binary mask maintaining target cardinality $K_i=\lfloor(1-S_i)\dth\rfloor$, with $S_i$ the sparsity ratio at stage $i$, and $\delta_i\in\R^\dth$ is a sparse, trainable delta.

The masks form a strictly nested sequence ($\supp(M_{i+1}) \subseteq \supp(M_i)$), preventing already pruned parameters from re-entering. The support $\Phi_i$ of the sparse delta is restricted strictly to the surviving coordinates ($\supp(\delta_{i}) \subseteq \Phi_i \subseteq \supp(M_i)$), with at most $\dph$ entries, where $\dph$ is fixed from the dense-model budget. We index the $L$ prunable weight matrices by $\ell$ and refer to them as modules.
Within each pruning step, \etaps performs restricted adaptation of the sparse delta under the current support, corresponding to the objective:
\begin{equation}
    \min_{\delta:
        \supp(\delta)\subseteq\Phi_i}
    \loss_{\mathcal D}
    \left(
        M_i \odot (\theta + \delta)
    \right).
    \label{eq:stage-adaptation}
\end{equation}
At initialization, \etaps processes $C$ calibration sequences in micro-batches to estimate the gradient statistic used to construct $M_1$ and initialize the delta support $\Phi_1$. Sparse-delta adaptation at stage $i$ compensates for past pruning perturbations while accumulating the gradient statistic used for the next pruning-mask selection in the same backward passes.

\subsection{Pruning Schedules and Adaptation Intervals}
\label{sec:schedule-saliency}

To schedule the pruning steps, \etaps uses the cubic gradual-pruning profile of \citet{zhu2018prune}:
\begin{equation}
  S_i
  =
  \Smax
  +
  (S_0-\Smax)
  \left(
      1-\frac{i}{P}
  \right)^3,
  \qquad
  i=1,\ldots,P,
  \label{eq:cubic}
\end{equation}
with $S_0=0$ as the starting point of the schedule and $S_P=\Smax$, which defines exactly $P$ pruning boundaries.

Separately from the sparsity profile, we allocate adaptation time non-uniformly: intervals $k_i$ between boundaries grow linearly with the stage index, assigning progressively longer adaptation intervals to later, higher-sparsity stages. This increasing allocation is an \etaps scheduling choice and is evaluated against a uniform allocation in \autoref{sec:ablation-allocation-main}. Over a total optimizer-step budget $T$, the cumulative step boundaries $b_i$ and interval durations $k_i$ are assigned as:
\begin{equation}
    b_i = \operatorname{round}\left[ T \frac{i(i+1)}{P(P+1)} \right], \quad b_0=0, \qquad k_i = b_i - b_{i-1}, \quad i = 1,\ldots,P.
    \label{eq:intervals}
\end{equation}
Pruning boundary $i$ is applied at optimizer step $b_{i-1}$ and followed by $k_i$ adaptation steps. With default $P=32$ and  $T=4096$, the interval grows from $k_1=8$ to $k_P=248$;  the final interval is therefore a fixed-mask recovery phase occupying \pc{6.1} of the optimization budget and does not accumulate statistics for another boundary.

\subsection{Gradient-Weighted Global Selection}
\label{sec:saliency}

At each stage, \etaps accumulates a gradient statistic from the same backward passes used for sparse-delta adaptation. Let $\mathcal B_{i,r}$ denote the $r$-th micro-batch processed during interval $i$, and let $\loss_{\mathcal B_{i,r}}$ denote its mean token loss. We accumulate
\begin{equation}
  \bar g_{i,j}
  =
  \frac{1}{n_i}
  \sum_{r=1}^{n_i}
  \left|
      \nabla_{w_{i,j}}
      \loss_{\mathcal B_{i,r}}(w_{i,r})
  \right|,
  \label{eq:grad_accum}
\end{equation}
where $n_i$ counts micro-batches (equivalently, backward passes), not sequences or optimizer steps. If each optimizer step accumulates gradients over $A$ micro-batches, then $n_i=k_iA$. The element-wise absolute value is taken on each micro-batch gradient before optimizer-gradient accumulation, so \autoref{eq:grad_accum} is the mean of absolute micro-batch gradients rather than the absolute value of an accumulated gradient. The same backward pass supplies the active-delta update and this saliency statistic, so no additional forward/backward pass is required; \etaps nevertheless retains the dense backward computation and incurs the element-wise accumulation and dense saliency-buffer cost. For the initial dense state, the $C$ calibration sequences are processed in micro-batches and $n_0$ is the resulting number of calibration backward passes.

At boundary $i+1$, the accumulated statistic is combined with the current effective-weight magnitude in a gradient-weighted score motivated by gradient-based pruning criteria \citep{lecun1989obd,lee2019snip,sanh2020movement,zhang2022platon}:
\begin{equation}
  s_{i,j}
  =
  \big|M_{i,j}(\theta_j + \delta_{i,j})\big|\,
  \bar g_{i,j},
  \qquad
  j\in\supp(M_i).
  \label{eq:saliency}
\end{equation}
Adaptation therefore changes both the boundary-time effective magnitudes and the interval-aggregated language-modeling gradients used to select the next pruning support. Because the magnitude factor is evaluated at the pruning boundary whereas $\bar g_{i,j}$ averages absolute gradients over the preceding interval, \autoref{eq:saliency} is an interval-aggregated saliency score rather than a single-state Taylor estimate. At boundary $i+1$, \etaps retains the globally highest-scoring $K_{i+1}$ surviving coordinates. We implement this selection with the four-pass streaming histogram and residual-ranking procedure in \autoref{alg:global-select}, without retaining a second dense score buffer, while enforcing the requested global cardinality.

\subsection{Sparse-Delta Support Reallocation}
\label{sec:relocation}

While the pruning support is strictly monotone ($\supp(M_{i+1})\subseteq\supp(M_i)$), the delta support $\Phi_i$ is relocated at pruning boundaries. Surviving delta entries retain their state, and vacated slots are reassigned within each module to  surviving coordinates with the largest accumulated gradient magnitude $\bar g_j$, following the gradient-guided growth principle used in dynamic sparse training \citep{evci2020rigl}.

Module delta budgets $B_\ell$ are fixed at initialization. If a module's survivor count falls below $B_\ell$, its active delta count contracts rather than reallocating the unused budget to other modules. Thus, the active delta count remains bounded by $\dph$: the pruning support monotonically removes coordinates, while the delta support redistributes only among surviving coordinates within each module.

\textbf{Relation to \seft.} \seft \citep{xiao2025seft} also uses a fixed-budget sparse delta whose support is periodically refreshed with gradient information from its fine-tuning objective. Its role is different: \seft starts from a model already pruned to the target sparsity, allows delta locations to move onto previously pruned coordinates, and then updates the pruning mask to restore the target sparsity. \etaps instead starts from the dense checkpoint and couples delta-support reallocation to a monotone pruning trajectory: reallocated delta entries are restricted to currently surviving coordinates, and the adapted effective model directly determines the next pruning support.

\section{Experiments}
\label{sec:experiments}

We evaluate \etaps on five model families: OPT-125M/1.3B \citep{zhang2022opt}, Gemma-2-2B/27B \citep{gemmateam2024gemma2improvingopen}, LLaMA-3.2-3B \citep{grattafiori2024llama3herdmodels}, LLaMA-2-7B/13B/70B \citep{touvron2023llama2}, and DeepSeek-67B \citep{deepseekai2024deepseekllm}. We report WikiText-2/C4 perplexity \citep{merity2017pointer,raffel2020c4} and zero-shot accuracy over seven downstream tasks using the Language Model Evaluation Harness \citep{eval-harness}. Comparators are \magnitude, \wanda \citep{sun2024wanda}, \sparsegpt \citep{frantar2023sparsegpt}, \alps \citep{meng2024alps}, and \elsa/\textsc{Elsa-L} \citep{lee2026elsa}; post-pruning recovery uses full fine-tuning, \lora \citep{hu2022lora}, and \seft \citep{xiao2025seft}; and semi-structured $N{:}M$ uses MaskLLM \citep{fang2024maskllm}. Trained methods use exactly $32{,}768$ length-$2048$ C4 sequence presentations; one-shot methods use 128 calibration sequences. Unless stated otherwise, we set the sparse delta budget $\dph$ to the number of trainable parameters that a rank-32 \lora would add over the same prunable module. Full protocol, baseline provenance, and reproduction details are given in \appref{app:exp_details}.

\subsection{Main Results}
\label{sec:main_results}

\autoref{tab:main-results} summarizes unstructured pruning on LLaMA-2-7B and LLaMA-2-13B at \pc{70}--\pc{90} sparsity. Across both models and all three sparsity levels, \etaps achieves the lowest WikiText-2/C4 perplexity and the highest mean zero-shot accuracy among
the evaluated methods. Results at moderate sparsity and on additional model
families are reported in \appref{app:full-results}.

\begin{table}[t]
    \centering
    \setlength{\tabcolsep}{4.5pt}
    \caption{
        LLaMA-2-7B and LLaMA-2-13B at \pc{70}--\pc{90} sparsity:
        perplexity ($\downarrow$) and average zero-shot accuracy ($\uparrow$, \%). The \elsa 13-B results are taken from \citep{lee2026elsa}.
    }
    \small
    \label{tab:main-results}
    \begin{tabular}{@{}clccc|ccc@{}}
        \toprule
        & &
        \multicolumn{3}{c}{\textbf{LLaMA-2-7B}} &
        \multicolumn{3}{c}{\textbf{LLaMA-2-13B}} \\
        \cmidrule(lr){3-5}
        \cmidrule(lr){6-8}
        \textbf{Sparsity} &
        \textbf{Method} &
        \textbf{Wiki} $\downarrow$ &
        \textbf{C4} $\downarrow$ &
        \textbf{Avg.} $\uparrow$ &
        \textbf{Wiki} $\downarrow$ &
        \textbf{C4} $\downarrow$ &
        \textbf{Avg.} $\uparrow$ \\
        \midrule

        \pc{0}
        & \dense
        & 5.47 & 7.26 & 60.21
        & 4.88 & 6.72 & 63.37 \\
        \midrule

        \multirow{6}{*}{\pc{70}}
        & \magnitude
        & 5.0e4 & 2.8e4 & 33.55
        & 218.1 & 196.4 & 33.91 \\

        & \wanda
        & 77.08 & 81.51 & 35.29
        & 45.19 & 55.31 & 38.44 \\

        & \sparsegpt
        & 27.77 & 31.37 & 41.70
        & 19.65 & 23.25 & 45.94 \\

        & \alps
        & 18.88 & 20.18 & 45.38
        & 14.13 & 16.35 & 49.44 \\

        & \elsa
        & 12.42 & 13.69 & 46.73
        & 11.14 & 12.20 & 50.30 \\

        & \textbf{\etaps}
        & \textbf{10.99} & \textbf{10.87} & \textbf{49.19}
        & \textbf{9.37} & \textbf{9.29} & \textbf{56.29} \\
        \midrule

        \multirow{6}{*}{\pc{80}}
        & \magnitude
        & 1.1e5 & 1.1e5 & 33.13
        & 3.9e4 & 4.8e4 & 32.89 \\

        & \wanda
        & 2226 & 2010 & 32.25
        & 1127 & 877.2 & 32.44 \\

        & \sparsegpt
        & 119.4 & 100.7 & 32.54
        & 94.64 & 78.75 & 34.65 \\

        & \alps
        & 48.20 & 47.21 & 36.85
        & 38.48 & 36.99 & 38.55 \\

        & \elsa
        & 17.99 & 17.34 & 41.77
        & 17.21 & 16.60 & 43.27 \\

        & \textbf{\etaps}
        & \textbf{14.17} & \textbf{13.09} & \textbf{44.50}
        & \textbf{11.93} & \textbf{11.10} & \textbf{50.20} \\
        \midrule

        \multirow{6}{*}{\pc{90}}
        & \magnitude
        & 4.8e4 & 5.4e4 & 32.84
        & 7.4e4 & 6.4e4 & 32.12 \\

        & \wanda
        & 6477 & 5099 & 32.62
        & 2.0e4 & 1.1e4 & 32.67 \\

        & \sparsegpt
        & 1369 & 1033 & 32.24
        & 1294 & 911.9 & 36.05 \\

        & \alps
        & 232.2 & 174.5 & 32.20
        & 241.2 & 149.8 & 32.92 \\

        & \elsa
        & 27.98 & 23.20 & 38.41
        & 30.19 & 25.07 & 37.37 \\

        & \textbf{\etaps}
        & \textbf{20.98} & \textbf{17.39} & \textbf{38.84}
        & \textbf{17.50} & \textbf{14.89} & \textbf{41.94} \\

        \bottomrule
    \end{tabular}
\end{table}

\subsection{Extreme Sparsity and Post-Pruning Recovery}
\label{sec:extreme}
 \autoref{tab:extreme} reports LLaMA-2-7B at \pc{90}, \pc{95}, and \pc{99} sparsity to examine behavior in the extreme-sparsity regime. In addition to \etaps and \elsa, we evaluate one-shot pruning with \wanda and \sparsegpt and post-pruning recovery applied after the target sparse mask has already been constructed. For \seft, we use both \wanda and \sparsegpt initializations, providing a direct test of how strongly recovery depends on the starting sparse checkpoint. 
\begin{table}[h!]
    \centering
    \small
    \caption{
        Extreme sparsity perplexity on LLaMA-2-7B.
    }
    \label{tab:extreme}
    \begin{tabular}{@{}llcccccc@{}}
        \toprule
        \multirow{2}{*}{\textbf{Model}} &
        \multirow{2}{*}{\textbf{Method}} &
        \multicolumn{2}{c}{\textbf{\pc{90}}} &
        \multicolumn{2}{c}{\textbf{\pc{95}}} &
        \multicolumn{2}{c}{\textbf{\pc{99}}} \\
        \cmidrule(lr){3-4}
        \cmidrule(lr){5-6}
        \cmidrule(lr){7-8}
        & & Wiki & C4 & Wiki & C4 & Wiki & C4 \\
        \midrule

        \multirow{8}{*}{
            \makecell[c]{\textbf{LLaMA-2-7B}\\
            \textit{Dense: 5.47 / 7.26}}
        }
        & \wanda
        & 6477 & 5099
        & 9558 & 7951
        & 1.24e5 & 1.44e5 \\

        & \sparsegpt
        & 1369 & 1033
        & 5075 & 5628
        & 1.51e4 & 9158 \\

        & \wanda + \lora
        & 43.67 & 34.33
        & 75.52 & 48.41
        & 249.6 & 91.25 \\

        & \wanda + Full
        & 31.27 & 24.30
        & 69.45 & 39.50
        & 195.75 & 68.62 \\

        & \wanda + \seft
        & 38.71 & 30.73
        & 79.49 & 47.37
        & 160.3 & 64.85 \\

        & \sparsegpt + \seft
        & 30.67 & 26.24
        & 50.81 & 37.04
        & 92.45 & 56.46 \\

        & \elsa
        & 27.98 & 23.20 
        & 33.22 & 28.14 
        & 55.58 & 39.98 \\

        & \textbf{\etaps}
        & \textbf{20.98} & \textbf{17.39}
        & \textbf{28.56} & \textbf{22.12}
        & \textbf{46.13} & \textbf{33.22} \\
        \bottomrule
    \end{tabular}
\end{table}
Post-pruning recovery improves one-shot models but remains sensitive to the initial sparse support. \etaps instead constructs the support during adaptation and achieves lower WikiText-2/C4 perplexity than all evaluated recovery methods and \elsa at \pc{90}--\pc{99} sparsity, with \pc{14.0}--\pc{25.0} lower perplexity than \elsa across the six dataset--sparsity comparisons. The matched progressive-versus-direct control in \autoref{sec:ablation-direct-main} separately isolates the effect of the pruning trajectory under the same adaptation budget.

\subsection{Execution at Large Model Scale}
\label{sec:Scalability}
We evaluate \etaps on models up to 70B parameters at \pc{90} sparsity. As shown in \autoref{tab:large_scale}, \etaps achieves 16.85/18.44 WikiText-2/C4 perplexity on Gemma-2-27B, compared with 28.47/29.27 for \textsc{Elsa-L} baseline. We further apply \etaps to DeepSeek-67B and LLaMA-2-70B, obtaining 13.00/14.65 and 11.69/11.00 perplexity, respectively. These results demonstrate that the same pruning procedure can be applied to models up to 70B parameters without changing its core formulation.
\begin{table}[ht]
    \centering
    \small
    \setlength{\tabcolsep}{6pt}
    \caption{
        Perplexity ($\downarrow$) at \pc{90} unstructured sparsity across 27B--70B parameter models. \elsa-L utilizes quantized optimizer and auxiliary state and is reproduced on Gemma-2-27B (its maximum reported scale); 67B and 70B settings are compared against scalable one-shot baselines.
    }
    \label{tab:large_scale}
    \begin{tabular}{@{}lcccccc@{}}
        \toprule
        & \multicolumn{2}{c}{\textbf{Gemma-2-27B}}
        & \multicolumn{2}{c}{\textbf{DeepSeek-67B}}
        & \multicolumn{2}{c}{\textbf{LLaMA-2-70B}} \\
        \cmidrule(lr){2-3} \cmidrule(lr){4-5} \cmidrule(lr){6-7}
        \textbf{Method} & Wiki & C4 & Wiki & C4 & Wiki & C4 \\
        \midrule
        \dense  & 5.68 & 9.26 & 7.11 & 11.06 & 3.32 & 5.71 \\
        \midrule
        Magnitude   & 4.6e12          & 4.2e12          & 5.1e5          & 5.0e5          & 1.5e5          & 9.0e4          \\
        \wanda       & 1.5e4           & 1.1e4           & 1.3e6          & 1.2e6          & 5319.8         & 4375.0         \\
        \sparsegpt  & 404.6           & 337.1           & 245.6          & 305.0          & 268.4          & 143.4          \\
        \elsa-L     & 28.47           & 29.27           & --             & --             & --             & --             \\
        \textbf{\etaps} & \textbf{16.85} & \textbf{18.44} & \textbf{13.00} & \textbf{14.65} & \textbf{11.69} & \textbf{11.00} \\
        \bottomrule
    \end{tabular}
\end{table}

We next examine pruning cost on LLaMA-2-7B at \pc{90} sparsity. \autoref{tab:cost} reports wall-clock time, aggregate GPU-hours, persistent pruning state, and final perplexity on NVIDIA H200 NVL GPUs. One-shot methods are substantially cheaper but degrade sharply at this sparsity, while post-pruning recovery improves quality at additional compute cost.

\begin{table}[h!]
    \centering
    \setlength{\tabcolsep}{4.2pt}
    \caption{Resource overhead and perplexity ($\downarrow$) for \pc{90}-sparse LLaMA-2-7B on NVIDIA H200 NVL GPUs. Persistent state denotes aggregate batch-independent model, optimizer, auxiliary, and saliency memory (excluding activations).}
    \label{tab:cost}
    \small
    \begin{tabular}{@{}lcccccc@{}}
        \toprule
        & \multicolumn{2}{c}{\textbf{Perplexity ($\downarrow$)}}
        & \multicolumn{3}{c}{\textbf{Resource Overhead}}
        & \multicolumn{1}{c}{\textbf{Memory (GiB)}} \\
        \cmidrule(lr){2-3} \cmidrule(lr){4-6} \cmidrule(l){7-7}
        \textbf{Method} & \textbf{Wiki} & \textbf{C4} & \textbf{GPUs} & \textbf{Wall-clock (h)} & \textbf{GPU-h} & \textbf{Persistent State} \\
        \midrule
        Magnitude      & 4.8e4               & 5.4e4               & 1 & \textbf{$<$0.01} & \textbf{$<$0.01} & 12.55 \\
        \wanda          & 6477                & 5099                & 1 & 0.01             & 0.01             & 12.55 \\
        \sparsegpt     & 1369                & 1033                & 1 & 0.06             & 0.06             & 12.55 \\
        \alps          & 232.2               & 174.5               & 1 & 0.92             & 0.92             & \textbf{4.04} \\
        \midrule
        \wanda + \seft  & 38.71               & 30.73               & 1 & 1.70             & 1.70             & 14.43 \\
        \wanda + \lora  & 43.67               & 34.33               & 1 & 2.32             & 2.32             & \textbf{12.91} \\
        \wanda + Full FT& \textbf{31.27}               & \textbf{24.30}               & 2 & 2.56             & 5.12             & 103.73 \\
        \midrule
        \elsa          & 27.98               & 23.20               & 4 & 0.81             & 3.24             & 132.29 \\
        \textbf{\etaps}& \textbf{20.98}      & \textbf{17.39}      & 1 & 1.66             & 1.66             & \textbf{26.40} \\
        \bottomrule
    \end{tabular}
\end{table}

\etaps reaches 20.98/17.39 WikiText-2/C4 perplexity in 1.66 GPU-hours on a single H200. In contrast, \elsa reproduction reaches 27.98/23.20 using 3.24 GPU-hours across four H200s, although its parallel execution gives a lower wall-clock time. \etaps therefore uses \pc{49} fewer aggregate GPU-hours while achieving lower perplexity in this setting. It also requires 26.40~GiB of aggregate persistent pruning state, compared with 132.29~GiB for \elsa, a $5.0\times$ reduction. The reported persistent-state measurements include model, optimizer, auxiliary, and saliency state but exclude activations and other transient buffers.

\subsection{Ablation Study}
\label{sec:ablation-main}
We examine four design choices using LLaMA-2-7B: gradual versus direct pruning, adaptation-interval schedules, saliency score formulations, and sparse-delta capacity. Additional model scale extensions (LLaMA-3.2-3B and OPT-1.3B) and results can be found in \appref{app:ablations}.

\subsubsection{One Step or Gradually?}
\label{sec:ablation-direct-main}
At moderate sparsity, direct one-shot pruning with sparse-delta repair performs similarly to progressive \etaps. However, the gap emerges at high sparsity (\autoref{tab:ablation-direct-main}). On LLaMA-2-7B at \pc{90} sparsity, direct projection degrades WikiText-2/C4 perplexity to 34.17/23.14, compared to 20.98/17.39 for \etaps. We attribute this to the size of the single pruning step: removing \pc{90} of the weights at once leaves a perturbation that the fixed-budget sparse delta cannot fully compensate, whereas progressive interleaving lets the delta adapt at each intermediate mask before the next pruning decision. Increasing from $P{=}1$ (direct) to $P{=}4$ recovers most of the degradation (34.17 to 22.04 WikiText-2), while further increases up to $P{=}32$ yield smaller gains (22.04$\to$20.98; \appref{app:ablation-boundaries}).

\begin{table}[ht]
    \centering
    \small
    \setlength{\tabcolsep}{6pt}
    \caption{Direct versus progressive pruning on LLaMA-2-7B (perplexity $\downarrow$).}
    \label{tab:ablation-direct-main}
    \begin{tabular}{@{}lcccccc@{}}
        \toprule
        & \multicolumn{2}{c}{\textbf{\pc{50}}}
        & \multicolumn{2}{c}{\textbf{\pc{70}}}
        & \multicolumn{2}{c}{\textbf{\pc{90}}} \\
        \cmidrule(lr){2-3}\cmidrule(lr){4-5}\cmidrule(lr){6-7}
        \textbf{Route} & Wiki & C4 & Wiki & C4 & Wiki & C4 \\
        \midrule
        Progressive (\etaps) & 6.66 & \textbf{8.20} & \textbf{10.99} & \textbf{10.87} & \textbf{20.98} & \textbf{17.39} \\
        Direct projection    & \textbf{6.54} & 8.23 & 11.47 & 11.34 & 34.17 & 23.14 \\
        \bottomrule
    \end{tabular}
\end{table}

\subsubsection{Uniform or Increasing Adaptation Intervals?}
\label{sec:ablation-allocation-main}

Under a fixed total optimizer budget, increasing and uniform adaptation intervals perform comparably at \pc{50} sparsity but diverge at higher levels. As shown in \autoref{tab:ablation-allocation-main}, an increasing interval allocation yields lower perplexity than a uniform one at \pc{90} sparsity.

\begin{table}[ht]
    \centering
    \small
    \setlength{\tabcolsep}{6pt}
    \caption{Adaptation-interval allocation on LLaMA-2-7B (perplexity $\downarrow$).}
    \label{tab:ablation-allocation-main}
    \begin{tabular}{@{}lcccccc@{}}
        \toprule
        & \multicolumn{2}{c}{\textbf{\pc{50}}}
        & \multicolumn{2}{c}{\textbf{\pc{70}}}
        & \multicolumn{2}{c}{\textbf{\pc{90}}} \\
        \cmidrule(lr){2-3}\cmidrule(lr){4-5}\cmidrule(lr){6-7}
        \textbf{Allocation} & Wiki & C4 & Wiki & C4 & Wiki & C4 \\
        \midrule
        Increasing (\etaps) & \textbf{6.66} & \textbf{8.20} & \textbf{10.99} & \textbf{10.87} & \textbf{20.98} & \textbf{17.39} \\
        Uniform             & 6.68          & 8.24          & 11.18          & 11.05          & 23.19          & 18.63          \\
        \bottomrule
    \end{tabular}
\end{table}

\subsubsection{Gradient-Weighted Saliency}
\label{sec:ablation-saliency-main}

\autoref{tab:ablation-saliency-main} compares three saliency criteria. While performance remains close at \pc{50} sparsity, alternative scores degrade rapidly at higher sparsities. At \pc{90} sparsity, magnitude-only selection increases WikiText-2/C4 perplexity to 32.64/24.52 (compared to 20.98/17.39 for the default mean absolute gradient), while taking the absolute value after batch-averaging signed gradients degrades performance further to 33.22/25.05.

\begin{table}[ht]
    \centering
    \small
    \setlength{\tabcolsep}{5pt}
    \caption{Saliency-score variants on LLaMA-2-7B (perplexity $\downarrow$).}
    \label{tab:ablation-saliency-main}
    \begin{tabular}{@{}lcccccc@{}}
        \toprule
        & \multicolumn{2}{c}{\textbf{\pc{50}}}
        & \multicolumn{2}{c}{\textbf{\pc{70}}}
        & \multicolumn{2}{c}{\textbf{\pc{90}}} \\
        \cmidrule(lr){2-3}\cmidrule(lr){4-5}\cmidrule(lr){6-7}
        \textbf{Score Variant} & Wiki & C4 & Wiki & C4 & Wiki & C4 \\
        \midrule
        Mean absolute gradient (\etaps) & \textbf{6.66} & \textbf{8.20} & \textbf{10.99} & \textbf{10.87} & \textbf{20.98} & \textbf{17.39} \\
        Magnitude only                  & 6.71          & 8.56          & 12.06          & 12.56          & 32.64          & 24.52          \\
        Absolute mean gradient          & 9.31          & 10.68         & 17.55          & 15.66          & 33.22          & 25.05          \\
        \bottomrule
    \end{tabular}
\end{table}

\subsubsection{Does the Sparse-Delta Budget Matter?}
\label{sec:ablation-budget-main}
Setting $\dph=0$ disables sparse-delta adaptation while keeping the progressive mask updates unchanged. At \pc{90}, this control collapses to 463.15/349.50 WikiText-2/C4 perplexity, whereas even a small positive budget ($r{=}8$) recovers to 20.70/18.80 on LLaMA-2-7B. Sparse adaptation is therefore essential in the high-sparsity regime.

We use $r{=}32$ as a fixed default rather than tuning the budget per model or sparsity level. Here $r$ denotes a \emph{LoRA-equivalent parameter budget}, not a low-rank \etaps parameterization: $\dph$ equals the number of trainable parameters that rank-$r$ \lora would add to the same prunable modules. Exact model-specific budgets are reported in \autoref{tab:rank-budget}, with the complete budget sweeps in \appref{app:ablation-budget}.

\section{Discussion}
\label{sec:discussion}

Our work addresses the challenge of preserving quality at extreme sparsity in LLM pruning, where inexpensive one-shot methods degrade sharply while methods that optimize the sparse mask directly from the dense model can require model-sized optimizer state. We propose \etaps, a progressive pruning framework in which gradients from the end-to-end loss guide pruning while optimizer state is maintained only for a sparse delta. Across our evaluated sparsity range, \etaps achieves lower perplexity than \elsa with substantially less persistent memory, and the same core procedure scales to models up to 70B parameters.

Our central finding is that gradual pruning and sparse adaptation are complementary. Without sparse adaptation, progressive pruning collapses at high sparsity. Without gradual pruning, the same sparse delta also falls short: under a matched adaptation budget, direct projection remains competitive at \pc{50}, but its gap to progressive pruning widens as the pruning target increases. The sweep over the number of pruning stages further suggests that this advantage comes mainly from moving beyond single-step pruning to multiple stages, rather than requiring a fine-grained schedule with many stages. Finally, the saliency controls show that gradient weighting becomes increasingly important in the high-sparsity regime.

\paragraph{Limitations.} While \etaps substantially reduces optimizer memory, it does not reduce the cost of backward computation: the backward pass remains dense, and the bf16 saliency accumulator is also dense. Thus, the memory savings come from restricting optimizer state to the sparse delta rather than from reducing backward FLOPs. The state metrics reported here also exclude activations and transient buffers, and therefore should not be interpreted as peak runtime VRAM. For larger models (27B--70B), we use multi-GPU execution primarily to accommodate memory capacity rather than to scale throughput; this demonstrates execution feasibility but does not establish an efficient multi-GPU implementation. 

Our experimental setup also involved limited hyperparameter tuning. Rather than independently tuning each model--sparsity configuration, we selected a single model-specific peak learning rate at \pc{90} sparsity and kept the other protocol choices fixed. The pruning support itself is strictly monotonic, and the reported results are based on single experimental runs. Moreover, the pruning schedule, delta budget, and module-wise sparsity allocation also remain largely empirical design choices, leaving room for more principled approaches. Finally, while perplexity and multiple-choice zero-shot accuracy are useful indicators of language modeling capability, they do not fully capture open-ended generation quality, reasoning capabilities, calibration, or safety preservation, nor do we evaluate end-to-end sparse inference latency.

\paragraph{Conclusion.} Our findings show that the current quality--cost trade-off curve at extreme sparsity is not a fixed frontier. Strong quality in this regime need not require the optimization-memory costs associated with adapting the full model. By combining gradual pruning with sparse adaptation, \etaps achieves comparable or better quality than \elsa while using only a fraction of the persistent optimization memory, suggesting that there is room to substantially reduce the cost of optimization at extreme sparsity without sacrificing model quality. Building on this result, future work can explore how the pruning trajectory and sparse adaptation can be designed more effectively, including how pruning stages are scheduled, how adaptation is allocated across stages, and how the delta budget is distributed across modules. We believe these findings provide a strong starting point for further advances in extreme-sparsity pruning and for exploring approaches that improve the balance between model quality and pruning cost in this challenging regime.

\bibliography{iclr2027_conference}
\bibliographystyle{iclr2027_conference}

\appendix
\section{Extended Related Work}
\label{app:related}
\paragraph{One-shot pruning with surrogate objectives.} One-shot pruning is computationally attractive because it produces a sparse model in a single step without retraining. Most existing methods achieve this by relying on a cheap surrogate objective to determine weight importance, following the approach pioneered by Optimal Brain Damage \citep{lecun1989obd}. \sparsegpt \citep{frantar2023sparsegpt} approximates layer-wise reconstruction error with a Hessian estimated from calibration activations, pruning weights while compensating updates to the survivors. \wanda \citep{sun2024wanda} uses a simpler saliency metric, weight magnitude $|W_{ij}|$ scaled by the norm $\|X_j\|_2$ of the corresponding input activation over the calibration set, without updating surviving weights. Pruner-Zero \citep{dong2024prunerzero} moves beyond manually designed importance scores by using symbolic regression to search for pruning criteria. $D^2$Prune \citep{xiong2026d2prune} addresses the mismatch between calibration and test-time activations by extending reconstruction-based pruning to jointly model weight and activation changes through a dual Taylor expansion. However, most one-shot pruners follow a local formulation, preventing comparisons across layers and requiring layer-wise sparsity to be specified separately. While most methods assume uniform ratios, OWL \citep{yin2024owl} instead derives non-uniform ratios from layer-wise weight-outlier distributions. We take a different approach by computing saliency directly from end-to-end loss gradients, allowing weights to compete across layers and sparsity allocation to emerge naturally from a single global ranking. This removes the need for a separate surrogate objective and predefined layer-wise sparsity ratios.

\paragraph{Iterative pruning with retraining.} Interleaving pruning with intermediate retraining has long been used to recover accuracy lost during pruning \citep{han2015learning}. More recently, \citet{janusz2025oneshot} systematically compare one-shot and iterative pruning on vision models and a small language model, and find one-shot preferable at low pruning ratios and iterative pruning ahead at high ones. Two qualifications of their result bear stating: under a fixed retraining budget they place the crossover near \pc{80}, and the regime they ultimately recommend is a hybrid that takes one large initial cut before switching to fine iterative steps. Their setting is neither LLM-scale nor post-training, so we read the ordering it establishes as motivation rather than as evidence for our own regime. At LLM scale, \citet{syed2023pruneandtune} similarly show that interleaving \sparsegpt pruning with short fine-tuning improves high-sparsity performance, while \citet{wip2025iterative} use iterative pruning with row-column weighted importance to prevent channel collapse and outperform one-shot pruning at high sparsity. Iterative pruning can follow different schedules, including progressively increasing sparsity toward the target \citep{zhu2018prune,hubens2022onecycle,kurtic2024smc} and alternating removal with regrowth to allow previously pruned weights to recover \citep{liu2021granet}. A related observation comes from sparse pre-training, where \citet{jin2025journey} find that evaluation loss is well modelled by the \emph{average} parameter count maintained over training rather than the final count, so that among schedules reaching the same target, those maintaining more parameters for longer are favoured. The optimum they report is interior rather than monotone: pruning begins at $25\%$ of the training compute and concludes at $75\%$. Their setting is pre-training rather than post-training pruning, and our schedule removes most weights early in optimizer-step time; we cite it only as evidence that the pruning trajectory, not just the final sparsity, affects final quality. The challenge, however, is applying this paradigm at a billion-parameter scale, where iterative pruning requires multiple rounds of recovery retraining over the model parameters \citep{syed2023pruneandtune}. \etaps addresses this problem by allocating sparse-delta and optimizer state for the small initial capacity $\dph$, set in our experiments to the \lora rank-$32$-equivalent parameter count. The active count $\dphi{i}\le\dph$ can shrink as layer supports contract, while inactive slots remain allocated along with their optimizer state.

\paragraph{Constrained optimization for pruning.} A different family formulates pruning as a constrained optimization problem, $\min_w \loss(w)$ s.t. $\|w\|_0\le K$, where the surviving weights emerge from optimization rather than from a separate scoring criterion. SparseFW \citep{roux2025sparsefw} relaxes the binary mask-selection problem to a convex $\ell_1$-constrained formulation and solves it with Frank--Wolfe, accounting for weight interactions while minimising per-layer reconstruction error. \alps \citep{meng2024alps} solves the layer-wise reconstruction problem under an $\ell_0$-constraint using ADMM, jointly optimizing the support and the surviving weights, while \citet{boza2024admm} applies ADMM to recover optimal weights on a fixed support. SAFE \citep{lee2025safe} extends the constraint to the whole model, directly optimizing a sparsity-constrained training objective through an augmented Lagrangian. \elsa applies surrogate-free ADMM to the model-level objective under a global sparsity constraint and provides empirical evidence that high-quality solutions can be reached at extreme sparsity levels well beyond those reached by conventional one-shot pruning \citep{lee2026elsa}. However, the computational cost of this method stems from its global formulation: the loss is optimized through dense backpropagation over all $\dth$ parameters, while introducing additional ADMM-based auxiliary variables in $\mathbb{R}^{\dth}$ to enforce the constraint. \etaps retains a model-level objective and exact cardinality constraints while reducing the allocated trainable optimization state: sparse-delta and optimizer storage is fixed at $\dph\ll\dth$ slots, of which only $\dphi{i}\le\dph$ are active between pruning boundaries. Dense effective-weight gradients are accumulated during adaptation for global saliency, and global $\TopK$ selection enforces the next cardinality exactly without maintaining model-sized ADMM dual variables throughout optimization.

\paragraph{Mask learning.} Mask learning treats the sparse support itself as a learnable object, directly optimizing mask variables rather than pruning weights via an importance criterion. MaskLLM \citep{fang2024maskllm} targets $N{:}M$ sparsity by parameterizing each group's valid patterns as a categorical distribution and using Gumbel-softmax to obtain a differentiable approximation to discrete pattern selection during training, and PATCH \citep{hourri2025patch} learns instead which tiles of a weight matrix are dense and which are $2{:}4$, so that its learnable state is one choice per tile rather than per weight, at the cost of an effective sparsity capped at \pc{50}. LEAP \citep{mozaffari2026leap} uses a per-weight Bernoulli with Gumbel-sigmoid relaxation, making end-to-end unstructured mask learning tractable at \pc{50}--\pc{60} sparsity. Except where the parameterization is coarsened as in PATCH, these methods optimize mask variables separately from the model weights, requiring auxiliary mask state on the order of the parameter count and dense per-step gradients in the per-weight case. We instead use gradient statistics to select the sparse support while adapting a much smaller sparse delta, so the state used for recovery also changes the next pruning decision.

\paragraph{Parameter-efficient sparse adaptation.} Parameter-efficient methods reduce the cost of adapting LLMs to downstream tasks by restricting the degrees of freedom available for fine-tuning. \lora \citep{hu2022lora} achieves efficiency by introducing a separate low-rank update instead of the full weight matrices, inspiring its sparse-native variants \citep{lu2024spp,munoz2024sqft,hu2026lors} to further constrain these updates to preserve the sparsity of the base model. However, the \lora formulation limits the expressiveness of updates to the chosen rank. Sparse fine-tuning methods take a different approach by concentrating weight adaptation in a sparse set of trainable coordinates \citep{guo2021diffpruning,ansell2022ltsft,ansell2024spiel}, where \seft \citep{xiao2025seft} is most relevant to our setting. \seft adapts an already pruned model with two coupled mechanisms: a delta support update that periodically reallocates which coordinates are trainable, including onto previously pruned ones, and a topology adaptation step that prunes the base model back to the target sparsity afterwards. \etaps reallocates a sparse trainable set in the same spirit, but couples it to progressive pruning, restricting reselection to surviving weights so that the target sparsity is never exceeded and no separate restoration step is needed.

\paragraph{Dynamic sparse training.} Dynamic sparse training (DST) studies the training problem where the model is initialized and updated with a sparse parameterization throughout. Its defining mechanism is that sparse connectivity evolves during training through an alternating pruning and regrowth mechanism. SET \citep{mocanu2018set} established this prune-and-regrow paradigm with random regrowth, while RigL \citep{evci2020rigl} uses dense gradient magnitude to guide regrowth. Top-KAST \citep{jayakumar2020topkast} maintains a separate exploration set of parameters to identify promising connections for future activations, while BiDST \citep{ji2024bidst} formulates prune-and-regrowth as a bi-level optimization problem. More recently, SMET \citep{xiao2026smet} studies the optimization instability caused by newly regrown parameters in sparse pre-training of LLaMA-style models from $60$M to $1$B, a characteristic challenge in DST, and addresses it with optimizer warm-up together with density-aware learning-rate scaling; we adopt the first of the two. \etaps draws on this observation when reallocating its trainable parameter budget: newly selected parameters have no optimizer history and therefore often receive excessively large updates. We similarly use a warm-up mechanism to stabilize their introduction during recovery. Beyond LLM finetuning, DST has also been successfully applied in model compression for deep networks with extremely large output layers which arise in scenarios of recommendation systems and tagging \citep{ullah2024navigating, ullah2026haste}.

\section{Detailed Algorithm}
\subsection{Overall Procedure}
	\label{sec:overall-procedure}

	\autoref{alg:main-algo} summarizes unstructured \etaps; in $N{:}M$ mode, the global $\TopK$ step is replaced by the group-wise projection in \autoref{alg:nm}.

	\begin{algorithm}[t]
		\caption{\etaps, unstructured mode.}
		\label{alg:main-algo}
		\begin{algorithmic}[1]
			\Require Pretrained weights $\theta$ (frozen); schedule parameters $S_0, \Smax$; delta budget $\dph$; boundaries $P$; total steps $T$; calibration count $C$
			\Ensure Sparse model state $w = M_P \odot (\theta + \delta_P)$
			\State Initialize $M_0 \equiv \mathbf 1$, $\Phi_0 = \varnothing$, $\delta_0 \gets \mathbf 0$
			\State Compute schedule $\{S_i\}_{i=1}^{P}$ via \eqref{eq:cubic} with $S_P = \Smax$
			\State Allocate intervals $\{k_i\}_{i=1}^{P}$ via \eqref{eq:intervals}
			\State Set $\widetilde B\gets\lfloor\dph/L\rfloor$; set $B_\ell\gets\min\{\widetilde B,d_\ell\}$ for each module and leave any remainder or capped shortfall unused
			\State Accumulate dense calibration gradient statistic $\bar g$ over $C$ sequences
			\For{$i = 1, \ldots, P$}
			\State Evaluate saliency $s_j \gets |M_{i-1,j}(\theta_j + \delta_{i-1,j})|\,\bar g_j$ for $j \in \supp(M_{i-1})$
			\State Compute target survivor count $K_i \gets \min\left(\lfloor(1-S_i)\dth\rfloor, \|M_{i-1}\|_0\right)$
			\State $M_i \gets$ the $K_i$ highest-scoring coordinates of $\supp(M_{i-1})$, ranked model-wide by streaming histogram selection (\autoref{sec:saliency})
			\State Deactivate delta entries pruned in $M_i$; carry each survivor's value, $(m,v)$, and local Adam step with its coordinate
			\For{each module $\ell$}
			\State Compute target active delta count $|\Phi_i^{(\ell)}| \gets \min\{B_\ell, \|M_i^{(\ell)}\|_0\}$
			\State Retain surviving active set $\Phi_{i-1}^{(\ell)}$ and refill up to $|\Phi_i^{(\ell)}|$ using largest $\bar g_j$ in module $\ell$
			\State For each new coordinate $j$: set $\delta_j\gets0$, $m_j\gets0$, $v_j\gets\bar g_j^2$, $t_j\gets1$, and apply its fixed local LR ramp
			\EndFor
			\State Reset gradient accumulator $\bar g \gets \mathbf 0$
			\State Compute active delta count $\dphi{i} \gets \sum_\ell |\Phi_i^{(\ell)}|$
			\For{$t = 1, \ldots, k_i$}
			\State Update active delta $\delta_{\Phi_i}$ with AdamW on $\loss_{\mathcal D}\left(M_i \odot (\theta + \delta)\right)$
			\If{$i < P$}
			\State Accumulate micro-batch gradient contributions into $\bar g$
			\EndIf
			\EndFor
			\EndFor
			\State \Return $w = M_P \odot (\theta + \delta_P)$
		\end{algorithmic}
	\end{algorithm}

    \begin{algorithm}[t]
        \caption{Global $\TopK$ selection.}
        \label{alg:global-select}
        \begin{algorithmic}[1]
            \Require Module scores $s^{(\ell)}$ on $\supp(M_{i-1}^{(\ell)})$, recomputed on demand for $\ell = 1,\ldots,L$; budget $K$; bins $B=4096$
            \Ensure Masks $M^{(\ell)}$ with $\sum_\ell \|M^{(\ell)}\|_0 = K$
            \State $n_+ \gets \sum_{\ell,j}\mathbf{1}[s_j^{(\ell)}>0]$ \Comment{pass 1}
            \If{$n_+\le K$}
                \State Keep all positive-score coordinates and fill the remaining $K-n_+$ slots from zero-score coordinates by stable global index order; \Return
            \EndIf
            \State $a \gets \log_2 \min_{\ell,j}\{s_j^{(\ell)}:s_j^{(\ell)}>0\}$, \quad $b \gets \log_2 \max_{\ell,j}s_j^{(\ell)}$
            \If{$a=b$}
                \State Retain exactly $K$ positive-score coordinates by stable global index order; \Return
            \EndIf
            \State $h \gets \sum_\ell \textsc{Histogram}\!\left(\{\log_2 s_j^{(\ell)}:s_j^{(\ell)}>0\};\, B\text{ bins on }[a,b]\right)$ \Comment{pass 2}
            \State $\beta \gets \min\left\{c:\sum_{c'\ge c}h_{c'}\le K\right\}$
            \State $\tau \gets 2^{a+\beta(b-a)/B}$, \quad $\tau_- \gets 2^{a+(\beta-1)(b-a)/B}$
            \State $M^{(\ell)} \gets \mathbf{1}[s^{(\ell)}\ge\tau]$ for all $\ell$ \Comment{pass 3}
            \State $r \gets K-\sum_\ell\|M^{(\ell)}\|_0$
            \If{$r>0$} \Comment{pass 4}
                \State Pool at most $r$ scores with $\tau_-\le s_j^{(\ell)}<\tau$ from each module
                \State Admit the $r$ highest pooled scores, breaking ties by stable global index order
            \ElsIf{$r<0$}
                \State Remove the $|r|$ lowest-scoring selected coordinates, breaking ties by stable global index order
            \EndIf
            \State \Return $\{M^{(\ell)}\}_{\ell=1}^{L}$
        \end{algorithmic}
    \end{algorithm}

	\subsection{Memory and Optimization State}
	\label{sec:memory}

	Only the sparse delta carries optimizer moments, so optimizer state is $\mathcal O(\dph)$; global selection additionally uses one dense bf16 $\mathcal O(\dth)$ gradient-statistic accumulator. Base parameters and the accumulator use bf16, while delta values and AdamW moments use fp32. We report pruning-state memory as model, optimizer, delta/index, and saliency storage that does not scale with the activation batch, excluding activations. During gradient accumulation it scales as:
	\begin{equation}
		\mathcal O(\dth)+\mathcal O(\dph),
		\label{eq:memory-state}
	\end{equation}
	with $\dph$ fixed for a given model throughout pruning and optimizer moments restricted exclusively to the $\mathcal O(\dph)$ component.

	The measured LLaMA-2-7B state values corresponding to this definition are reported in \autoref{tab:cost}. They are pruning-state-memory measurements rather than peak runtime-memory measurements.

\section{Experimental Details}
	\label{app:exp_details}

	\subsection{Implementation and Reproduction Details}
	The implementation uses PyTorch \citep{paszke2019pytorch} and draws primarily on the sparse-adaptation implementation of \seft \citep{xiao2025seft}. All author-run experiments use NVIDIA H200 NVL GPUs. \etaps uses one GPU through LLaMA-2-13B, two for Gemma-2-27B, and four for DeepSeek-67B and LLaMA-2-70B. The current multi-GPU implementation is used primarily to fit larger models in memory rather than to improve throughput: some phases execute on a single GPU even when multiple devices are allocated, and phases that use two GPUs leave any additional devices idle. These runs show execution at the reported model sizes but are not a distributed-scaling study.
    
     Unless otherwise noted, Magnitude, \wanda, \sparsegpt, and \alps are
    reproduced across the unstructured model grid, while \lora, full
    fine-tuning, \seft, and unstructured \elsa are reproduced through 7B.
    For LLaMA-2-7B, we reproduce \elsa from \pc{50} through \pc{99} using
    the authors' official implementation and recommended hyperparameters;
    our reproduced values are used throughout. At \pc{90}, \pc{95}, and
    \pc{99}, our runs give 27.98/23.20, 33.22/28.14, and 55.58/39.98
    WikiText-2/C4 perplexity, respectively. LLaMA-2-13B \elsa values are
    taken from \citet{lee2026elsa}, and Gemma-2-27B \elsa-L is reproduced.
        
    For $N{:}M$, MaskLLM $2{:}4$ is evaluated from the released checkpoint under the same evaluation protocol.
    
    Reproduced baselines use held-out data only when tuning is required; recommended configurations are otherwise used directly. Hyperparameter-search budgets are not assumed to be identical across methods, and reported WikiText-2/C4 test values are never used for hyperparameter selection.
    
    \subsection{Training and Evaluation Data}

	Calibration and adaptation use C4 \citep{raffel2020c4}. Following the \elsa data protocol, \etaps uses 128 sequences for initial calibration and $32{,}768$ length-$2048$ C4 sequence presentations for adaptation at every target sparsity (about $67.1$M tokens); the optimizer-step and batch-size settings that realize this fixed budget are listed in \autoref{tab:hyperparams}. Reproduced \elsa and trained recovery controls use the corresponding C4 adaptation budget, while one-shot baselines use 128 calibration sequences; literature-sourced rows retain their original protocols. The trained and one-shot rows are therefore lower-cost calibration references rather than data-matched training competitors. Zero-shot accuracy is reported on ARC-Challenge, ARC-Easy, BoolQ, HellaSwag, OpenBookQA, RTE, and Winogrande using one fixed Language Model Evaluation Harness configuration, with the mean taken over these seven tasks. The exact harness version, task metric settings, software dependencies, and model revisions are pinned in the released reproducibility configuration.

	\subsection{Hyperparameters}
	\label{app:hyperparams}

	\begin{table}[ht]
		\centering
		\caption{\etaps optimization protocol by target-sparsity regime.}
		\label{tab:hyperparams}
		\small
		\setlength{\tabcolsep}{5.0pt}
		\begin{tabular}{@{}lcc@{}}
			\toprule
			\textbf{Setting} & \textbf{$\Smax\!\le\!\pc{60}$} & \textbf{$\Smax\!>\!\pc{60}$} \\
			\midrule
			Initial sparsity $S_0$ & 0 & 0 \\
			Pruning boundaries $P$ & 8 & 32 \\
			Optimizer steps $T$ & 1024$^\dagger$ & 4096 \\
			Effective batch size & 32$^\dagger$ & 8 \\
			\midrule
			Sparse-delta budget $\dph$ & \multicolumn{2}{c}{\lora rank-$32$-equivalent over prunable modules} \\
			Gradient-statistic accumulation & \multicolumn{2}{c}{full adaptation interval $k_i$} \\
			Optimizer & \multicolumn{2}{c}{AdamW; $\beta_1{=}0.9$, $\beta_2{=}0.999$, $\epsilon{=}10^{-6}$; fp32 moments} \\
			Weight decay & \multicolumn{2}{c}{0} \\
			Learning-rate schedule & \multicolumn{2}{c}{cosine decay} \\
			Run seed & \multicolumn{2}{c}{42} \\
			\bottomrule
		\end{tabular}

		\smallskip
		\footnotesize $^\dagger$OPT-125M uses $T{=}2048$ and effective batch size 16.
	\end{table}

	\begin{table}[ht]
		\centering
		\caption{Peak learning rates by model.}
		\label{tab:model-lr}
		\small
		\setlength{\tabcolsep}{7pt}
		\begin{tabularx}{0.92\linewidth}{@{}lX@{}}
			\toprule
			\textbf{Peak learning rate} & \textbf{Models} \\
			\midrule
			$1\mathrm{e}{-3}$ & OPT-125M; LLaMA-2-7B; Gemma-2-27B; DeepSeek-67B; LLaMA-2-70B \\
			$5\mathrm{e}{-4}$ & LLaMA-3.2-3B; LLaMA-2-13B \\
			$1\mathrm{e}{-4}$ & OPT-1.3B; Gemma-2-2B \\
			\bottomrule
		\end{tabularx}
	\end{table}

	For models evaluated at multiple sparsity levels, the peak learning rate is selected once at \pc{90} on a held-out C4 validation split, disjoint from adaptation and reported test data, and is then fixed across sparsity levels. The 27B--70B runs use $1\mathrm{e}{-3}$. %Delta-slot allocation and reallocation state are defined in \autoref{sec:method-overview} and \autoref{sec:relocation}, rather than treated as tunable hyperparameters.
We parameterize the nominal sparse-delta budget with a LoRA-equivalent rank $r$ only to obtain a consistent model-dependent parameter count. For the $L$ prunable matrices, $\dph(r)=r\sum_{\ell=1}^{L}(d_{\mathrm{out},\ell}+d_{\mathrm{in},\ell})$, equal to the number of trainable parameters that rank-$r$ \lora would add to those matrices. \etaps itself stores $\dph$ individual sparse coordinates rather than low-rank factors. The default is $r{=}32$. Module budgets satisfy $\sum_\ell B_\ell\le\dph$, and the active count $\dphi{i}$ can be smaller still as surviving supports contract (\autoref{sec:relocation}).

\begin{table}[ht]
    \centering\small
    \setlength{\tabcolsep}{5pt}
    \caption{Prunable parameter count $d_\theta$ and nominal LoRA-equivalent sparse-delta budget $\dph(r)$. A fixed $r$ is not a fixed fraction of the model: for the default $r{=}32$, $\dph/d_\theta$ decreases from \pc{6.25} on OPT-125M to \pc{0.61} on LLaMA-2-70B.}
    \label{tab:rank-budget}
    \begin{tabular}{@{}lrrrrr@{}}
        \toprule
        \multirow{2}{*}{\textbf{Model}} & \multirow{2}{*}{$d_\theta$}
        & \multicolumn{4}{c}{$\dph$} \\
        \cmidrule(lr){3-6}
        & & $r{=}8$ & $r{=}16$ & $r{=}32$ & $r{=}64$ \\
        \midrule
        OPT-125M     &     84,934,656 &   1,327,104 &   2,654,208 &   5,308,416 &  10,616,832 \\
        OPT-1.3B     &  1,207,959,552 &   7,077,888 &  14,155,776 &  28,311,552 &  56,623,104 \\
        Gemma-2-2B   &  2,024,275,968 &  10,383,360 &  20,766,720 &  41,533,440 &  83,066,880 \\
        LLaMA-3.2-3B &  2,818,572,288 &  12,156,928 &  24,313,856 &  48,627,712 &  97,255,424 \\
        LLaMA-2-7B   &  6,476,005,376 &  19,988,480 &  39,976,960 &  79,953,920 & 159,907,840 \\
        LLaMA-2-13B  & 12,687,769,600 &  31,293,440 &  62,586,880 & 125,173,760 & 250,347,520 \\
        Gemma-2-27B  & 26,046,627,840 &  57,090,048 & 114,180,096 & 228,360,192 & 456,720,384 \\
        DeepSeek-67B & 65,745,715,200 & 107,786,240 & 215,572,480 & 431,144,960 & 862,289,920 \\
        LLaMA-2-70B  & 68,451,041,280 & 103,546,880 & 207,093,760 & 414,187,520 & 828,375,040 \\
        \bottomrule
    \end{tabular}
\end{table}

\section{Semi-Structured \texorpdfstring{$N{:}M$}{N:M} Sparsity}
\label{sec:nm-pattern}

	For $N{:}M$ sparsity, \etaps replaces model-wide projection with a group-wise operator while keeping the same schedule, saliency, and sparse-delta adaptation. Weight matrices are partitioned into contiguous groups of $M$ parameters with at most $N$ survivors. $2{:}4$ matches the sparsity pattern supported by NVIDIA Sparse Tensor Cores; $4{:}8$ is evaluated only as a less restrictive \pc{50}-density pattern \citep{mishra2021nm}.

	An $N{:}M$ pattern fixes target sparsity to
	\begin{equation}
		\Smax
		=
		1-\frac{N}{M}.
		\label{eq:nm-target}
	\end{equation}
	The schedule thus governs the sequence of intermediate sparsity levels leading to this endpoint. At stage $i$, the scheduled average number of survivors per group is
	\begin{equation}
		\kappa_i
		=
		\operatorname{clip}\!\left(
		M(1-S_i),\,N,\,M
		\right).
		\label{eq:nm-kappa}
	\end{equation}
	Because individual groups must retain an integer number of parameters, we bracket $\kappa_i$ by
	\begin{equation}
		k_{\mathrm{lo}}
		=
		\max\!\left(\lfloor\kappa_i\rfloor,N\right),
		\qquad
		k_{\mathrm{hi}}
		=
		\min\!\left(\lceil\kappa_i\rceil,M\right).
		\label{eq:nm-levels}
	\end{equation}
	For non-integer $\kappa_i$, eligible groups receive either $k_{\mathrm{lo}}$ or $k_{\mathrm{hi}}$ survivors to match the scheduled average; previously tightened groups remain fixed. Live coordinates are ranked by the same saliency as unstructured \etaps, with a fixed deterministic ordering used only to choose which eligible groups transition. The constraint therefore tightens monotonically:
	\begin{equation}
		M{:}M
		\;\longrightarrow\;
		\cdots
		\;\longrightarrow\;
		N{:}M.
	\end{equation}
	\autoref{alg:nm} details group-wise mask selection.
	\begin{algorithm}[t]
		\caption{Progressive $N{:}M$ mask selection at boundary $i$.}
		\label{alg:nm}
		\begin{algorithmic}[1]
			\Function{StagedNM}{$\mathbf{s}, M, S_i, N, E$}
			\Require Saliency scores $\mathbf{s}$; group size $M$; scheduled sparsity $S_i$; final survivors $N$; current support $E$
			\Ensure Monotone group-wise mask
			\State $\kappa_i \gets \operatorname{clip}(M(1-S_i),N,M)$
			\State $k_{\mathrm{lo}} \gets \max(\lfloor\kappa_i\rfloor,N)$, $k_{\mathrm{hi}} \gets \min(\lceil\kappa_i\rceil,M)$
			\State Reshape $\mathbf{s}$ and $E$ into $G$ contiguous groups; mask coordinates outside $E$
			\State Compute $t_i$, the number of eligible groups assigned $k_{\mathrm{lo}}$ survivors to match the scheduled average
			\State If $k_{\mathrm{lo}}<k_{\mathrm{hi}}$, assign these transitions using the fixed boundary-saliency ordering
			\State Retain the highest-saliency live coordinates up to the assigned survivor count in each group
			\State \Return Updated mask
			\EndFunction
		\end{algorithmic}
	\end{algorithm}

	At boundary $P$, $S_P = 1-N/M$ and every group reaches the target $N{:}M$ pattern. \autoref{tab:nm-pattern} reports the LLaMA-2-7B semi-structured evaluation. MaskLLM is evaluated from the released $2{:}4$ checkpoint under the same standard evaluation protocol; no $4{:}8$ MaskLLM checkpoint is included.

\begin{table}[ht]
    \centering
    \setlength{\tabcolsep}{3.2pt}
    \caption{LLaMA-2-7B semi-structured sparsity: perplexity and seven-task zero-shot accuracy.}
    \label{tab:nm-pattern}
    \resizebox{\linewidth}{!}{%
    \begin{tabular}{@{}clcc|cccccccc@{}}
        \toprule
        & & \multicolumn{2}{c}{\textbf{Perplexity} ($\downarrow$)} & \multicolumn{8}{c}{\textbf{Tasks} ($\uparrow$)} \\
        \cmidrule(lr){3-4}\cmidrule(lr){5-12}
        \textbf{Pattern} & \textbf{Method} & \textbf{Wiki} & \textbf{C4} & \textbf{ARC-C} & \textbf{ARC-E} & \textbf{BoolQ} & \textbf{HellaSwag} & \textbf{OBQA} & \textbf{RTE} & \textbf{Winogrande} & \textbf{Avg.} \\
        \midrule
        -- & \dense & 5.47 & 7.26 & 42.75 & 75.55 & 79.33 & 57.13 & 33.40 & 63.54 & 69.77 & 60.21 \\
        \midrule
        $2{:}4$ & MaskLLM & \textbf{7.33} & \textbf{9.13} & 35.24 & 69.23 & 71.47 & 51.28 & 28.40 & 57.76 & 64.48 & \textbf{53.98} \\
        $2{:}4$ & \etaps & 7.50 & 9.41 & 33.62 & 67.59 & 71.83 & 50.40 & 28.20 & 59.57 & 64.96 & 53.74 \\
        $4{:}8$ & \etaps & 7.01 & 8.89 & 33.79 & 67.68 & 74.43 & 51.70 & 28.40 & 65.34 & 66.06 & 55.34 \\
        \bottomrule
    \end{tabular}}
\end{table}

\section{Additional Ablations and Diagnostics}
\label{app:ablations}

The main text reports the central mechanistic controls on LLaMA-2-7B. This appendix extends those controls to LLaMA-3.2-3B and OPT-1.3B and reports the complete boundary, schedule, budget, allocation, and mask diagnostics.

\subsection{Global versus Uniform Module Allocation}
\label{sec:ablation-layer-allocation}

Global selection jointly determines surviving weights and module-wise sparsity allocation. The resulting module density,
\begin{equation}
    \rho_{i,\ell} = \frac{\|M_i^{(\ell)}\|_0}{d_\ell},
    \label{eq:emergent-layer-density}
\end{equation}
emerges from global ranking. A matched \textsc{Uniform per-module} control holds the schedule, saliency, delta budget, and delta-support reallocation rule fixed. \autoref{tab:emergent-density} favors global allocation on C4 in all ten comparisons, while WikiText-2 is mixed; we therefore treat allocation as an empirical design choice rather than a universal advantage.

\begin{table}[ht]
    \centering
    \small
    \setlength{\tabcolsep}{3pt}
    \caption{Global versus uniform module-wise sparsity allocation (perplexity $\downarrow$).}
    \label{tab:emergent-density}
        \begin{tabular}{@{}llcccccccccc@{}}
        \toprule
        \multirow{2}{*}{\textbf{Model}} &
        \multirow{2}{*}{\textbf{Allocation}} &
        \multicolumn{2}{c}{\textbf{\pc{50}}} &
        \multicolumn{2}{c}{\textbf{\pc{60}}} &
        \multicolumn{2}{c}{\textbf{\pc{70}}} &
        \multicolumn{2}{c}{\textbf{\pc{80}}} &
        \multicolumn{2}{c}{\textbf{\pc{90}}} \\
        \cmidrule(lr){3-4} \cmidrule(lr){5-6} \cmidrule(lr){7-8} \cmidrule(lr){9-10} \cmidrule(lr){11-12}
        & & Wiki & C4 & Wiki & C4 & Wiki & C4 & Wiki & C4 & Wiki & C4 \\
        \midrule
        \multirow{2}{*}{LLaMA-3.2-3B}
        & Uniform & \textbf{10.00} & 14.16 & \textbf{12.55} & 16.76 & 17.51 & 20.45 & 24.51 & 26.03 & 37.39 & 35.46 \\
        & Global & 10.14 & \textbf{13.97} & 12.85 & \textbf{16.61} & \textbf{17.12} & \textbf{19.93} & \textbf{23.82} & \textbf{25.48} & \textbf{36.01} & \textbf{35.15} \\
        \midrule
        \multirow{2}{*}{LLaMA-2-7B}
        & Uniform & \textbf{6.58} & 8.30 & \textbf{7.53} & 9.17 & 11.33 & 11.22 & 15.25 & 13.49 & \textbf{20.88} & 17.78 \\
        & Global & 6.66 & \textbf{8.20} & 7.69 & \textbf{9.04} & \textbf{10.99} & \textbf{10.87} & \textbf{14.17} & \textbf{13.09} & 20.98 & \textbf{17.39} \\
        \bottomrule
    \end{tabular}
\end{table}

\par
Module-density profiles in \autoref{tab:layer-density-profile} become more non-uniform as sparsity rises (CV: $0.32\to0.86$ for LLaMA-3.2-3B and $0.46$ to $0.98$ for LLaMA-2-7B). These profiles are descriptive; the matched global-versus-uniform comparison provides the relevant ablation evidence.

\subsection{Direct-Projection Sweep}
\label{sec:ablation-direct}

\autoref{tab:ablation-direct} extends the main LLaMA-2-7B route comparison to LLaMA-3.2-3B and OPT-1.3B and additionally tests delta-support reallocation after direct projection on LLaMA-2-7B. Direct variants prune to the target during calibration and use the same sparse-delta optimization budget as the corresponding progressive run.

\begin{table}[ht]
    \centering\small
    \caption{Direct versus progressive pruning across three models (perplexity $\downarrow$).}
    \label{tab:ablation-direct}
    \begin{tabular}{@{}llcccccc@{}}
        \toprule
        & & \multicolumn{2}{c}{\textbf{\pc{50}}} & \multicolumn{2}{c}{\textbf{\pc{70}}} & \multicolumn{2}{c}{\textbf{\pc{90}}} \\
        \cmidrule(lr){3-4}\cmidrule(lr){5-6}\cmidrule(lr){7-8}
        \textbf{Model} & \textbf{Route} & Wiki & C4 & Wiki & C4 & Wiki & C4 \\
        \midrule
        \multirow{3}{*}{LLaMA-2-7B}
        & Progressive (\etaps)  & 6.66 & \textbf{8.20} & \textbf{10.99} & \textbf{10.87} & \textbf{20.98} & \textbf{17.39} \\
        & Direct projection     & 6.54 & 8.23 & 11.47 & 11.34 & 34.17 & 23.14 \\
        & Direct $+$ delta reallocation & \textbf{6.53} & 8.23 & 11.34 & 11.33 & 35.81 & 23.12 \\
        \midrule
        \multirow{2}{*}{LLaMA-3.2-3B}
        & Progressive (\etaps)  & \textbf{10.14} & \textbf{13.97} & \textbf{17.12} & \textbf{19.93} & \textbf{36.01} & \textbf{35.15} \\
        & Direct projection     & 10.67 & 14.73 & 19.84 & 22.39 & 60.40 & 49.97 \\
        \midrule
        \multirow{2}{*}{OPT-1.3B}
        & Progressive (\etaps)  & \textbf{14.84} & \textbf{16.65} & \textbf{20.26} & \textbf{20.33} & \textbf{54.64} & \textbf{42.30} \\
        & Direct projection     & 15.28 & 17.00 & 23.16 & 22.40 & 154.07 & 80.73 \\
        \bottomrule
    \end{tabular}
\end{table}

At \pc{90}, the progressive route is lower on both datasets for all three models. Delta-support reallocation after direct projection does not close the LLaMA-2-7B gap.

\subsection{Number of Pruning Boundaries}
\label{app:ablation-boundaries}

The largest gain occurs when moving away from one-shot projection: increasing $P$ from $1$ to $4$ reduces WikiText-2 perplexity by \pc{35.5}, whereas increasing $P$ from $8$ to $32$ yields a further \pc{4.7} reduction. C4 shows the same diminishing-return pattern. Because the total optimizer-step budget $T$ is fixed, this ablation varies boundary count without adding adaptation steps; it motivates the fixed choice of $P$ by sparsity band but does not imply that larger $P$ is universally preferable.
\begin{table}[ht]
    \centering\small
    \caption{Boundary-count and start-sparsity controls on LLaMA-2-7B at \pc{90} (perplexity $\downarrow$).}
    \label{tab:ablation-boundaries}
    \begin{tabular}{@{}lccccc@{}}
        \toprule
        $P$ & 1 (one-shot) & 4 & 8 & 16 & 32 (default) \\
        \midrule
        Wiki & 34.17 & 22.04 & 22.02 & 21.91 & \textbf{20.98} \\
        C4   & 23.14 & 18.12 & 17.68 & 17.45 & \textbf{17.39} \\
        \midrule
        $S_0$ & \pc{0} & \pc{25} & \pc{50} & \pc{70} & \pc{90} \\
        \midrule
        Wiki & 20.98 & 21.13 & \textbf{20.61} & 22.48 & 34.17 \\
        C4   & 17.39 & 17.27 & \textbf{17.18} & 17.45 & 23.14 \\
        \bottomrule
    \end{tabular}
\end{table}

The lower block varies the ramp origin $S_0$ at fixed $P$ and fixed $T$, shortening the sequence of intermediate sparsity levels without reducing the optimization budget. Starting at \pc{25} or \pc{50} changes perplexity only slightly relative to the dense start, while $S_0{=}\pc{70}$ raises WikiText-2 perplexity from $20.98$ to $22.48$ and $S_0{=}\pc{90}$ reduces to the direct-projection control ($34.17/23.14$). Thus, a fully dense start is not required in this experiment, but sufficiently late entry degrades quality. Because every row retains the same $T$, this ablation does \emph{not} demonstrate that beginning from a pre-sparse checkpoint permits the corresponding fraction of the optimization budget to be skipped.

\subsection{Adaptation-Interval Allocation}
\label{sec:ablation-allocation}

\autoref{tab:ablation-allocation} extends the main increasing-versus-uniform interval-allocation control to LLaMA-3.2-3B at the same total optimization budget.

\begin{table}[ht]
    \centering\small
    \caption{Adaptation-interval allocation on LLaMA-3.2-3B (perplexity $\downarrow$).}
    \label{tab:ablation-allocation}
    \begin{tabular}{@{}lcccccc@{}}
        \toprule
        & \multicolumn{2}{c}{\textbf{\pc{50}}} & \multicolumn{2}{c}{\textbf{\pc{70}}} & \multicolumn{2}{c}{\textbf{\pc{90}}} \\
        \cmidrule(lr){2-3}\cmidrule(lr){4-5}\cmidrule(lr){6-7}
        \textbf{Step allocation} & Wiki & C4 & Wiki & C4 & Wiki & C4 \\
        \midrule
        Increasing (\etaps) & \textbf{10.14} & \textbf{13.97} & \textbf{17.12} & \textbf{19.93} & \textbf{36.01} & \textbf{35.15} \\
        Uniform             & 10.17 & 14.04 & 17.38 & 20.48 & 39.66 & 37.84 \\
        \bottomrule
    \end{tabular}
\end{table}

The separation is small at \pc{50} and larger at \pc{90}, matching the trend in the LLaMA-2-7B main-text control.

\subsection{Schedule Shape}
\label{sec:ablation-schedule}

The matched increasing-versus-uniform interval-allocation control is reported in \autoref{tab:ablation-allocation-main} and \autoref{tab:ablation-allocation}. Here we instead vary the sparsity trajectory while holding the total step budget $T$, number of boundaries $P$, and step-allocation rule fixed. The cubic schedule in \eqref{eq:cubic} interpolates between $S_0$ and $\Smax$ with a cubic profile, so this comparison changes only the sparsity attained at each boundary. The cubic profile reaches \pc{87.5} of the total sparsity increase by the midpoint, so its later pruning increments are smaller than those of a linear profile; the geometric profile lies between the two. \autoref{tab:ablation-schedule} compares all three profiles on LLaMA-2-7B, LLaMA-3.2-3B, and OPT-1.3B.

The cubic profile is best in 17 of the 18 model--dataset--sparsity comparisons; the exception is LLaMA-2-7B WikiText-2 at \pc{50}, where linear is lower by only $0.01$ perplexity ($6.65$ vs. $6.66$). At \pc{50}, every geometric/linear result is within \pc{1.5} of the cubic result across all three models, whereas the separation is much larger at \pc{90}. There, linear raises WikiText-2 perplexity by \pc{44.5} on LLaMA-3.2-3B and \pc{89.5} on OPT-1.3B relative to cubic; geometric raises it by \pc{12.2} and \pc{19.4}, respectively. Because $T$, $P$, and the step-allocation rule are matched, the comparison changes the sparsity trajectory while holding total adaptation steps and boundary count fixed. It supports the cubic profile for the tested configurations, but does not identify a universal optimal schedule or show that the alternatives receive less total recovery compute.

\begin{table}[ht]
    \centering\small
    \setlength{\tabcolsep}{4pt}
    \caption{Sparsity schedule at fixed step and boundary budgets (perplexity $\downarrow$).}
    \label{tab:ablation-schedule}
    \begin{tabular}{@{}llcccccc@{}}
        \toprule
        & & \multicolumn{2}{c}{\textbf{\pc{50}}} & \multicolumn{2}{c}{\textbf{\pc{70}}} & \multicolumn{2}{c}{\textbf{\pc{90}}} \\
        \cmidrule(lr){3-4}\cmidrule(lr){5-6}\cmidrule(lr){7-8}
        \textbf{Model} & \textbf{Profile} & Wiki & C4 & Wiki & C4 & Wiki & C4 \\
        \midrule
        \multirow{3}{*}{LLaMA-2-7B}
        & Cubic (\etaps) & 6.66 & \textbf{8.20} & \textbf{10.99} & \textbf{10.87} & \textbf{20.98} & \textbf{17.39} \\
        & Geometric      & 6.72 & 8.25 & 11.40 & 11.22 & 23.98 & 18.91 \\
        & Linear         & \textbf{6.65} & 8.27 & 12.14 & 11.52 & 29.64 & 23.15 \\
        \midrule
        \multirow{3}{*}{LLaMA-3.2-3B}
        & Cubic (\etaps) & \textbf{10.14} & \textbf{13.97} & \textbf{17.12} & \textbf{19.93} & \textbf{36.01} & \textbf{35.15} \\
        & Geometric      & 10.18 & 14.14 & 18.09 & 21.01 & 40.41 & 38.32 \\
        & Linear         & 10.18 & 14.18 & 19.14 & 21.90 & 52.05 & 48.72 \\
        \midrule
        \multirow{3}{*}{OPT-1.3B}
        & Cubic (\etaps) & \textbf{14.84} & \textbf{16.65} & \textbf{20.26} & \textbf{20.33} & \textbf{54.64} & \textbf{42.30} \\
        & Geometric      & 15.02 & 16.69 & 21.58 & 21.18 & 65.24 & 48.37 \\
        & Linear         & 15.01 & 16.71 & 22.31 & 21.85 & 103.53 & 66.51 \\
        \bottomrule
    \end{tabular}
\end{table}

\subsection{Saliency Score}
\label{sec:ablation-saliency}

\autoref{tab:ablation-saliency} extends the LLaMA-2-7B main-text control to LLaMA-3.2-3B and OPT-1.3B. \textsc{Magnitude only} drops the gradient factor, while \textsc{Absolute mean gradient} takes the absolute value only after averaging signed interval gradients; all other settings are held fixed.

\begin{table}[ht]
    \centering\small
    \setlength{\tabcolsep}{5pt}
    \caption{Cross-model saliency-score extension (perplexity $\downarrow$).}
    \label{tab:ablation-saliency}
    \begin{tabular}{@{}llcccccc@{}}
        \toprule
        \textbf{Model} & \textbf{Score variant} & \multicolumn{2}{c}{\textbf{\pc{50}}} & \multicolumn{2}{c}{\textbf{\pc{70}}} & \multicolumn{2}{c}{\textbf{\pc{90}}} \\
        \cmidrule(lr){3-4}\cmidrule(lr){5-6}\cmidrule(lr){7-8}
        & & Wiki & C4 & Wiki & C4 & Wiki & C4 \\
        \midrule
        \multirow{3}{*}{LLaMA-3.2-3B}
        & Mean absolute gradient & \textbf{10.14} & \textbf{13.97} & \textbf{17.12} & \textbf{19.93} & \textbf{36.01} & \textbf{35.15} \\
        & Magnitude only         & 11.20 & 15.56 & 22.87 & 24.57 & 58.04 & 47.77 \\
        & Absolute mean gradient & 17.20 & 21.60 & 32.25 & 32.30 & 58.27 & 48.50 \\
        \midrule
        \multirow{3}{*}{OPT-1.3B}
        & Mean absolute gradient & \textbf{14.84} & \textbf{16.65} & \textbf{20.26} & \textbf{20.33} & \textbf{54.64} & \textbf{42.30} \\
        & Magnitude only         & 17.62 & 17.70 & 24.50 & 22.97 & 118.31 & 69.91 \\
        & Absolute mean gradient & 20.37 & 20.64 & 35.58 & 30.84 & 103.91 & 67.42 \\
        \bottomrule
    \end{tabular}
\end{table}

The mean-absolute-gradient score is lower than both alternatives in every reported LLaMA-3.2-3B and OPT-1.3B dataset--sparsity cell.

\subsection{Sparse-Delta Budget}
\label{app:ablation-budget}
Here $r$ indexes the LoRA-equivalent nominal budget defined in \autoref{tab:rank-budget}; it does not denote a low-rank \etaps update. For the two models below, $\dph=r\times1{,}519{,}616$ for LLaMA-3.2-3B and $\dph=r\times2{,}498{,}560$ for LLaMA-2-7B. The $r{=}0$ rows disable adaptation while retaining progressive mask updates.

\begin{table}[h!]
    \centering\small
    \setlength{\tabcolsep}{4pt}
    \caption{LLaMA-3.2-3B sparse-delta budget sweep (perplexity $\downarrow$; cells report WikiText-2 / C4).}
    \label{tab:ablation-budget-3b}
    \begin{tabular}{@{}lccccc@{}}
        \toprule
        \textbf{Budget} & \textbf{\pc{50}} & \textbf{\pc{70}} & \textbf{\pc{90}} & \textbf{\pc{95}} & \textbf{\pc{99}} \\
        \midrule
        $r{=}0$ (no adaptation) & 14.62 / 20.27 & 53.41 / 67.00 & 1783 / 1230 & 2.1e4 / 1.1e4 & 1.5e4 / 9710 \\
        \midrule
        $r{=}8$  & 10.38 / 14.20 & 17.73 / 20.93 & 42.08 / 41.23 & 62.32 / 56.26 & 122.2 / 95.04 \\
        $r{=}16$ & 10.25 / 14.05 & 17.24 / 20.35 & 38.31 / 37.46 & 55.34 / 48.87 & 100.6 / 77.41 \\
        $r{=}32$ (default) & 10.14 / 13.97 & 17.12 / 19.93 & 36.01 / 35.15 & 47.63 / 43.01 & 86.28 / 66.78 \\
        $r{=}64$ & \textbf{10.10 / 13.82} & \textbf{16.73 / 19.54} & \textbf{32.95 / 31.82} & \textbf{43.40 / 38.87} & \textbf{77.07 / 60.13} \\
        \bottomrule
    \end{tabular}
\end{table}

\begin{table}[h!]
    \centering\small
    \setlength{\tabcolsep}{4pt}
    \caption{LLaMA-2-7B sparse-delta budget sweep (perplexity $\downarrow$; cells report WikiText-2 / C4).}
    \label{tab:ablation-budget-7b}
    \begin{tabular}{@{}lccccc@{}}
        \toprule
        \textbf{Budget} & \textbf{\pc{50}} & \textbf{\pc{70}} & \textbf{\pc{90}} & \textbf{\pc{95}} & \textbf{\pc{99}} \\
        \midrule
        $r{=}0$ (no adaptation) & 8.31 / 10.60 & 22.52 / 25.92 & 463.2 / 349.5 & 5438 / 2243 & 8889 / 9852 \\
        \midrule
        $r{=}8$  & \textbf{6.62} / 8.20 & \textbf{10.19 / 10.73} & 20.70 / 18.80 & 31.29 / 25.46 & 69.39 / 44.15 \\
        $r{=}16$ & \textbf{6.62 / 8.18} & 10.46 / 10.74 & \textbf{20.67} / 18.01 & 30.88 / 23.56 & 56.61 / 37.10 \\
        $r{=}32$ (default) & 6.66 / 8.20 & 10.99 / 10.87 & 20.98 / 17.39 & 28.56 / 22.12 & 46.13 / 33.22 \\
        $r{=}64$ & 6.70 / 8.25 & 12.14 / 11.14 & 21.75 / \textbf{16.98} & \textbf{27.71 / 20.81} & \textbf{42.29 / 29.74} \\
        \bottomrule
    \end{tabular}
\end{table}

On LLaMA-3.2-3B, larger positive budgets consistently reduce perplexity from \pc{50} through \pc{99}. LLaMA-2-7B is less monotonic through \pc{90}, but larger budgets become increasingly helpful at \pc{95} and \pc{99}. We keep $r{=}32$ fixed in the primary experiments rather than tuning budget separately for each model--sparsity pair.

\subsection{Sparse-Delta Budget Utilization}
\label{app:offset-activation}

% Active vs deactivated delta slots (LLaMA-2-7B).
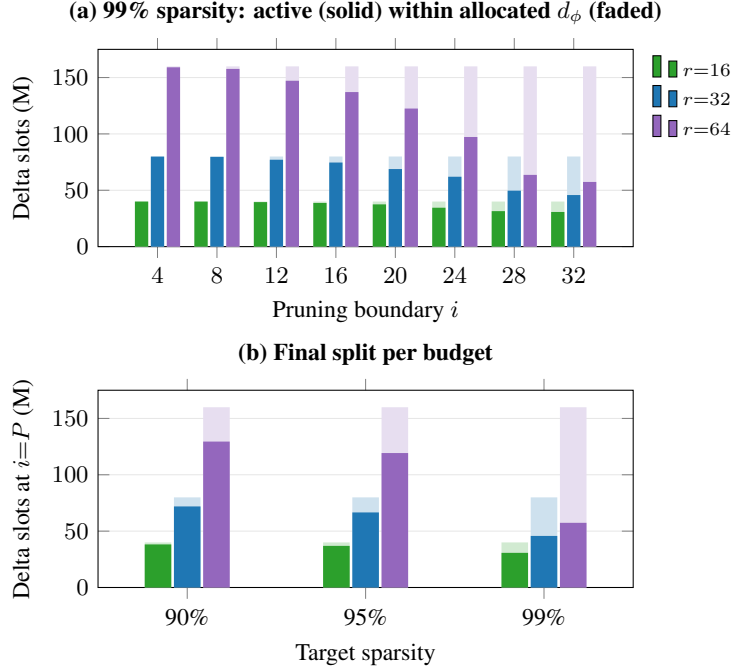
\begin{figure}[H]
\centering
\definecolor{c16}{HTML}{2CA02C}
\definecolor{c32}{HTML}{1F77B4}
\definecolor{c64}{HTML}{9467BD}
\begin{tikzpicture}
\begin{axis}[name=axL, width=0.62\textwidth, height=0.30\textwidth,
  ybar, bar width=5pt,
  xlabel={Pruning boundary $i$}, ylabel={Delta slots (M)},
  xlabel style={font=\small}, ylabel style={font=\small},
  tick label style={font=\footnotesize}, xmin=0, xmax=36, ymin=0, ymax=175,
  xtick={4,8,12,16,20,24,28,32},
  ymajorgrids, grid style={line width=0.25pt,draw=gray!20},
  title={(a) \pc{99} sparsity: active (solid) within allocated $\dph$ (faded)},
  title style={font=\small\bfseries},
  legend style={font=\scriptsize, at={(1.02,1.0)}, anchor=north west, draw=none,
                fill=none, cells={anchor=west}}]
\addplot[fill=c16!22, draw=none, bar shift=-6pt, forget plot] coordinates {(4,39.98) (8,39.98) (12,39.98) (16,39.98) (20,39.98) (24,39.98) (28,39.98) (32,39.98)};
\addplot[fill=c16, draw=none, bar shift=-6pt] coordinates {(4,39.98) (8,39.96) (12,39.49) (16,38.74) (20,37.48) (24,34.51) (28,31.47) (32,30.74)};
\addlegendentry{$r{=}16$}

\addplot[fill=c32!22, draw=none, bar shift=0pt, forget plot] coordinates {(4,79.95) (8,79.95) (12,79.95) (16,79.95) (20,79.95) (24,79.95) (28,79.95) (32,79.95)};
\addplot[fill=c32, draw=none, bar shift=0pt] coordinates {(4,79.87) (8,79.63) (12,77.08) (16,74.60) (20,68.82) (24,62.00) (28,49.66) (32,45.75)};
\addlegendentry{$r{=}32$}

\addplot[fill=c64!22, draw=none, bar shift=6pt, forget plot] coordinates {(4,159.91) (8,159.91) (12,159.91) (16,159.91) (20,159.91) (24,159.91) (28,159.91) (32,159.91)};
\addplot[fill=c64, draw=none, bar shift=6pt] coordinates {(4,159.07) (8,157.66) (12,147.15) (16,137.11) (20,122.49) (24,97.26) (28,63.64) (32,57.38)};
\addlegendentry{$r{=}64$}
\end{axis}

\begin{axis}[name=axR, at={($(axL.south)-(0,1.9cm)$)}, anchor=north,
  width=0.62\textwidth, height=0.30\textwidth,
  ybar, bar width=10pt,
  xlabel={Target sparsity}, ylabel={Delta slots at $i{=}P$ (M)},
  xlabel style={font=\small}, ylabel style={font=\small},
  tick label style={font=\footnotesize}, xmin=0.5, xmax=3.5, ymin=0, ymax=175,
  xtick={1,2,3}, xticklabels={\pc{90}, \pc{95}, \pc{99}},
  ymajorgrids, grid style={line width=0.25pt,draw=gray!20},
  title={(b) Final split per budget}, title style={font=\small\bfseries}]
\addplot[fill=c16!22, draw=none, bar shift=-11pt, forget plot] coordinates {(1,39.98) (2,39.98) (3,39.98)};
\addplot[fill=c16, draw=none, bar shift=-11pt, forget plot] coordinates {(1,38.21) (2,36.93) (3,30.74)};

\addplot[fill=c32!22, draw=none, bar shift=0pt, forget plot] coordinates {(1,79.95) (2,79.95) (3,79.95)};
\addplot[fill=c32, draw=none, bar shift=0pt, forget plot] coordinates {(1,71.91) (2,66.59) (3,45.75)};

\addplot[fill=c64!22, draw=none, bar shift=11pt, forget plot] coordinates {(1,159.91) (2,159.91) (3,159.91)};
\addplot[fill=c64, draw=none, bar shift=11pt, forget plot] coordinates {(1,129.46) (2,119.28) (3,57.38)};
\end{axis}
\end{tikzpicture}
\caption{Allocated delta slots (faded) and realized active capacity (solid, $\dphi{i}$) on LLaMA-2-7B; the gap corresponds to inactive capacity. Per-layer budgets $B_\ell$ are fixed at initialization, so slots retire as the surviving pruning support contracts. (a) Evolution across pruning boundaries at \pc{99} sparsity for three budgets. (b) Final active capacity at \pc{90}, \pc{95}, and \pc{99}.}
\label{fig:offset-utilisation}
\end{figure}

Because per-module budgets $B_\ell$ are fixed at initialization while module supports contract during pruning, not every allocated delta slot remains usable. When $\|M_i^{(\ell)}\|_0<B_\ell$, excess slots become inactive rather than being reallocated, so the realized active count $\dphi{i}$ can fall below the nominal budget $\dph$.

\autoref{fig:offset-utilisation} quantifies this effect on LLaMA-2-7B. At the final boundary, the inactive share of the default $r{=}32$ budget increases from \pc{10.1} at \pc{90} to \pc{16.7} at \pc{95} and \pc{42.8} at \pc{99}. At \pc{99}, increasing the nominal budget to $r{=}64$ raises the active count from $45.75$M to $57.38$M slots while increasing the inactive share to \pc{64.1}. Thus, nominal and realizable delta capacity diverge at extreme sparsity, helping explain the budget sweep in \autoref{tab:ablation-budget-7b}.

\subsection{Module Density Profile}

\autoref{tab:layer-density-profile} summarizes the final module-wise density distributions produced by global selection under the default configuration. Across all six models, module densities become increasingly heterogeneous as sparsity rises from \pc{50} to \pc{90}, while the identities of the most and least preserved modules remain architecture-dependent. These statistics are descriptive rather than causal; the matched global-versus-uniform comparison in \autoref{tab:emergent-density} provides the controlled evidence for the allocation rule. Mean density is parameter-weighted and fixed by the target sparsity, whereas min, median, max, and CV are unweighted across prunable modules.
\begin{table}[ht!]
    \centering
    \small
    \caption{Final module-density statistics under global selection.}
    \label{tab:layer-density-profile}
    \resizebox{\linewidth}{!}{%
        \begin{tabular}{@{}lcccccccc@{}}
            \toprule
            \textbf{Model} &
            \textbf{Sparsity} &
            \textbf{Mean} &
            \textbf{Min} &
            \textbf{Median} &
            \textbf{Max} &
            \textbf{CV} &
            \multicolumn{2}{c}{\textbf{Most / least preserved module}} \\
            \midrule

            \multirow{3}{*}{OPT-125M}
            & \pc{50} & \pc{50} & \pc{18.5} & \pc{50.8} & \pc{75.0} & 0.23
            & \texttt{L0.fc1} & \texttt{L3.fc2} \\
            & \pc{70} & \pc{30} & \pc{8.4} & \pc{28.0} & \pc{62.2} & 0.35
            & \texttt{L0.fc1} & \texttt{L3.fc2} \\
            & \pc{90} & \pc{10} & \pc{1.8} & \pc{8.9} & \pc{34.1} & 0.60
            & \texttt{L0.fc1} & \texttt{L3.k\_proj} \\
            \midrule

            \multirow{3}{*}{OPT-1.3B}
            & \pc{50} & \pc{50} & \pc{0.9} & \pc{52.6} & \pc{78.3} & 0.40
            & \texttt{L23.k\_proj} & \texttt{L1.q\_proj} \\
            & \pc{70} & \pc{30} & \pc{0.7} & \pc{26.4} & \pc{55.0} & 0.52
            & \texttt{L13.v\_proj} & \texttt{L1.q\_proj} \\
            & \pc{90} & \pc{10} & \pc{0.4} & \pc{8.3} & \pc{37.9} & 0.63
            & \texttt{L0.fc1} & \texttt{L1.k\_proj} \\
            \midrule

            \multirow{3}{*}{Gemma-2-2B}
            & \pc{50} & \pc{50} & \pc{12.1} & \pc{51.2} & \pc{83.2} & 0.32
            & \texttt{L7.v\_proj} & \texttt{L25.q\_proj} \\
            & \pc{70} & \pc{30} & \pc{4.0} & \pc{29.4} & \pc{72.4} & 0.50
            & \texttt{L7.v\_proj} & \texttt{L25.q\_proj} \\
            & \pc{90} & \pc{10} & \pc{0.8} & \pc{10.2} & \pc{61.8} & 0.95
            & \texttt{L4.v\_proj} & \texttt{L10.gate\_proj} \\
            \midrule

            \multirow{3}{*}{LLaMA-3.2-3B}
            & \pc{50} & \pc{50} & \pc{5.8} & \pc{50.8} & \pc{75.1} & 0.32
            & \texttt{L3.v\_proj} & \texttt{L0.q\_proj} \\
            & \pc{70} & \pc{30} & \pc{3.1} & \pc{27.1} & \pc{66.3} & 0.53
            & \texttt{L3.v\_proj} & \texttt{L25.q\_proj} \\
            & \pc{90} & \pc{10} & \pc{1.1} & \pc{8.3} & \pc{51.0} & 0.86
            & \texttt{L14.v\_proj} & \texttt{L9.gate\_proj} \\
            \midrule

            \multirow{3}{*}{LLaMA-2-7B}
            & \pc{50} & \pc{50} & \pc{0.7} & \pc{53.3} & \pc{71.5} & 0.46
            & \texttt{L2.down\_proj} & \texttt{L0.q\_proj} \\
            & \pc{70} & \pc{30} & \pc{0.3} & \pc{27.8} & \pc{59.3} & 0.66
            & \texttt{L2.down\_proj} & \texttt{L30.k\_proj} \\
            & \pc{90} & \pc{10} & \pc{0.1} & \pc{6.5} & \pc{42.0} & 0.98
            & \texttt{L2.down\_proj} & \texttt{L30.k\_proj} \\
            \midrule

            \multirow{3}{*}{LLaMA-2-13B}
            & \pc{50} & \pc{50} & \pc{0.6} & \pc{51.3} & \pc{70.3} & 0.49
            & \texttt{L2.down\_proj} & \texttt{L0.q\_proj} \\
            & \pc{70} & \pc{30} & \pc{0.3} & \pc{28.5} & \pc{60.6} & 0.71
            & \texttt{L2.down\_proj} & \texttt{L33.k\_proj} \\
            & \pc{90} & \pc{10} & \pc{0.1} & \pc{6.0} & \pc{44.6} & 1.00
            & \texttt{L2.down\_proj} & \texttt{L33.k\_proj} \\
            \bottomrule
        \end{tabular}%
    }
\end{table}

\FloatBarrier
\section{Complete Model Evaluation}
\label{app:full-results}
\subsection{LLaMA-2-7B task-wise results}
The task-wise values corresponding to the LLaMA-2-7B columns of \autoref{tab:main-results} are reported below.
\begin{table}[!ht]
    \centering
    \setlength{\tabcolsep}{3.2pt}
    \caption{LLaMA-2-7B perplexities and task-wise results at \pc{70}--\pc{90} sparsity.}
    \label{tab:main-results-llama-2-7b-full}
    \resizebox{\linewidth}{!}{%
        \begin{tabular}{@{}clcc|cccccccc@{}}
            \toprule
            & &
            \multicolumn{2}{c}{\textbf{Perplexity} ($\downarrow$)} &
            \multicolumn{8}{c}{\textbf{Tasks} ($\uparrow$)} \\
            \cmidrule(lr){3-4}
            \cmidrule(lr){5-12}
            \textbf{Sparsity} &
            \textbf{Method} &
            \textbf{Wiki} &
            \textbf{C4} &
            \textbf{ARC-C} &
            \textbf{ARC-E} &
            \textbf{BoolQ} &
            \textbf{HellaSwag} &
            \textbf{OBQA} &
            \textbf{RTE} &
            \textbf{Winogrande} &
            \textbf{Avg.} \\
            \midrule

            \pc{0} &
            \dense & 5.47 & 7.26 &
            42.75 & 75.55 & 79.33 & 57.13 &
            33.40 & 63.54 & 69.77 & 60.21 \\
            \midrule

            \multirow{6}{*}{\pc{70}}
            & \magnitude
            & 5.0e4 & 2.8e4
            & 24.23 & 28.58 & 38.10 & 25.91
            & 15.40 & 53.43 & 49.17 & 33.55 \\

            & \wanda
            & 77.08 & 81.51
            & 18.43 & 31.02 & 53.61 & 28.04
            & 12.60 & 52.71 & 50.59 & 35.29 \\

            & \sparsegpt
            & 27.77 & 31.37
            & 21.25 & 44.61 & 63.73 & 32.95
            & 18.20 & 53.07 & 58.09 & 41.70 \\

            & \alps
            & 18.88 & 20.18
            & 24.74 & 53.32 & 64.34 & 37.85
            & 21.80 & 53.43 & 62.19 & 45.38 \\

            & \elsa
            & 12.42 & 13.69
            & 27.65 & 56.23 & 63.91 & 43.53
            & 23.80 & 54.15 & 57.85 & 46.73 \\

            & \textbf{\etaps}
            & \textbf{10.99} & \textbf{10.87}
            & 30.12 & 61.74 & 66.79 & 46.49
            & 25.60 & 53.07 & 60.54 & 49.19 \\
            \midrule

            \multirow{6}{*}{\pc{80}}
            & \magnitude
            & 1.1e5 & 1.1e5
            & 20.14 & 25.84 & 46.33 & 25.64
            & 14.40 & 51.99 & 47.59 & 33.13 \\

            & \wanda
            & 2226 & 2010
            & 21.42 & 26.73 & 37.83 & 25.91
            & 12.40 & 53.43 & 48.07 & 32.25 \\

            & \sparsegpt
            & 119.4 & 100.7
            & 17.49 & 27.57 & 39.57 & 27.48
            & 12.60 & 54.15 & 48.93 & 32.54 \\

            & \alps
            & 48.20 & 47.21
            & 19.45 & 31.99 & 61.90 & 29.20
            & 12.00 & 51.70 & 51.70 & 36.85 \\

            & \elsa
            & 17.99 & 17.34
            & 20.99 & 47.18 & 62.11 & 36.38
            & 18.60 & 52.71 & 54.46 & 41.77 \\

            & \textbf{\etaps}
            & \textbf{14.17} & \textbf{13.09}
            & 22.70 & 54.84 & 61.99 & 40.55
            & 23.60 & 52.71 & 55.09 & 44.50 \\
            \midrule

            \multirow{6}{*}{\pc{90}}
            & \magnitude
            & 4.8e4 & 5.4e4
            & 23.55 & 26.01 & 42.48 & 25.65
            & 14.40 & 46.93 & 50.83 & 32.84 \\

            & \wanda
            & 6477 & 5099
            & 21.50 & 25.76 & 37.83 & 25.79
            & 14.80 & 52.71 & 49.96 & 32.62 \\

            & \sparsegpt
            & 1369 & 1033
            & 20.99 & 25.21 & 37.83 & 25.71
            & 14.60 & 52.71 & 48.62 & 32.24 \\

            & \alps
            & 232.2 & 174.5
            & 19.71 & 27.02 & 37.86 & 26.75
            & 13.40 & 54.51 & 46.17 & 32.20 \\

            & \elsa
            & 27.98 & 23.20
            & 19.80 & 40.53 & 57.49 & 31.10
            & 15.80 & 52.71 & 51.46 & 38.41 \\

            & \textbf{\etaps}
            & \textbf{20.98} & \textbf{17.39}
            & 19.97 & 43.98 & 55.23 & 32.72
            & 17.00 & 52.71 & 50.28 & 38.84 \\

            \bottomrule
        \end{tabular}%
    }
\end{table}

\subsection{LLaMA-2-13B task-wise results}
The task-wise values corresponding to the LLaMA-2-13B columns of \autoref{tab:main-results} are reported below.

\begin{table}[!ht]
    \centering
    \setlength{\tabcolsep}{3.2pt}
    \caption{LLaMA-2-13B perplexities and task-wise results at \pc{70}--\pc{90} sparsity. \elsa resultes are taken from \citep{lee2026elsa}.}
    \label{tab:main-results-llama-2-13b}
    \resizebox{\linewidth}{!}{%
        \begin{tabular}{@{}clcc|cccccccc@{}}
            \toprule
            & &
            \multicolumn{2}{c}{\textbf{Perplexity} ($\downarrow$)} &
            \multicolumn{8}{c}{\textbf{Tasks} ($\uparrow$)} \\
            \cmidrule(lr){3-4}
            \cmidrule(lr){5-12}
            \textbf{Sparsity} &
            \textbf{Method} &
            \textbf{Wiki} &
            \textbf{C4} &
            \textbf{ARC-C} &
            \textbf{ARC-E} &
            \textbf{BoolQ} &
            \textbf{HellaSwag} &
            \textbf{OBQA} &
            \textbf{RTE} &
            \textbf{Winogrande} &
            \textbf{Avg.} \\
            \midrule

            \pc{0} &
            \dense & 4.88 & 6.72 &
            47.10 & 78.91 & 82.35 & 60.21 &
            34.60 & 68.23 & 72.22 & 63.37 \\
            \midrule

            \multirow{6}{*}{\pc{70}}
            & \magnitude
            & 218.1 & 196.4
            & 20.39 & 31.69 & 38.17 & 27.75
            & 16.20 & 52.71 & 50.43 & 33.91 \\

            & \wanda
            & 45.19 & 55.31
            & 18.52 & 38.47 & 62.23 & 29.51
            & 13.80 & 52.71 & 53.83 & 38.44 \\

            & \sparsegpt
            & 19.65 & 23.25
            & 26.79 & 54.00 & 66.73 & 36.91
            & 20.60 & 52.71 & 63.85 & 45.94 \\

            & \alps
            & 14.13 & 16.35
            & 29.10 & 59.68 & 71.25 & 41.11
            & 25.20 & 54.15 & 65.59 & 49.44 \\

            & \elsa
            & 11.14 & 12.20
            & 34.13 & 62.42 & 70.12 & 47.52
            & 24.80 & 52.71 & 60.38 & 50.30 \\

            & \textbf{\etaps}
            & \textbf{9.37} & \textbf{9.29}
            & 36.52 & 69.57 & 74.34 & 53.25
            & 31.20 & 64.26 & 64.88 & \textbf{56.29} \\
            \midrule

            \multirow{6}{*}{\pc{80}}
            & \magnitude
            & 3.9e4 & 4.8e4
            & 22.10 & 25.55 & 41.41 & 25.78
            & 14.60 & 52.71 & 48.07 & 32.89 \\

            & \wanda
            & 1127 & 877.2
            & 20.39 & 26.56 & 37.83 & 26.45
            & 14.20 & 52.71 & 48.93 & 32.44 \\

            & \sparsegpt
            & 94.64 & 78.75
            & 18.43 & 27.86 & 54.40 & 27.86
            & 12.40 & 52.71 & 48.86 & 34.65 \\

            & \alps
            & 38.48 & 36.99
            & 19.97 & 35.86 & 62.45 & 30.28
            & 13.80 & 52.71 & 54.78 & 38.55 \\

            & \elsa
            & 17.21 & 16.60
            & 24.32 & 50.97 & 63.52 & 38.03
            & 19.60 & 52.71 & 53.75 & 43.27 \\

            & \textbf{\etaps}
            & \textbf{11.93} & \textbf{11.10}
            & 30.80 & 63.47 & 69.54 & 47.53
            & 26.60 & 52.71 & 60.77 & \textbf{50.20} \\
            \midrule

            \multirow{6}{*}{\pc{90}}
            & \magnitude
            & 7.4e4 & 6.4e4
            & 21.33 & 25.55 & 42.42 & 25.62
            & 14.80 & 45.49 & 49.64 & 32.12 \\

            & \wanda
            & 2.0e4 & 1.1e4
            & 20.90 & 25.76 & 37.83 & 26.03
            & 14.00 & 52.71 & 51.46 & 32.67 \\

            & \sparsegpt
            & 1294 & 911.9
            & 21.16 & 26.05 & 61.41 & 25.85
            & 14.40 & 52.71 & 50.75 & 36.05 \\

            & \alps
            & 241.2 & 149.8
            & 19.45 & 27.10 & 39.11 & 27.22
            & 13.80 & 53.43 & 50.36 & 32.92 \\

            & \elsa
            & 30.19 & 25.07
            & 19.03 & 36.15 & 58.44 & 28.65
            & 16.20 & 52.71 & 50.43 & 37.37 \\

            & \textbf{\etaps}
            & \textbf{17.50} & \textbf{14.89}
            & 22.53 & 49.87 & 54.92 & 38.04
            & 22.40 & 52.71 & 53.12 & \textbf{41.94} \\

            \bottomrule
        \end{tabular}%
    }
\end{table}

\subsection{Cross-Model Perplexity Sweep}

    \begin{table}[H]
    \centering
    \footnotesize
    \setlength{\tabcolsep}{4pt}
    \caption{Perplexity across model families at \pc{50}--\pc{90} sparsity.}
    \label{tab:main-ppl-results}
    \resizebox{\linewidth}{!}{%
        \begin{tabular}{@{}llcccccccccc@{}}
            \toprule
            \multirow{2}{*}{\textbf{Model}} &
            \multirow{2}{*}{\textbf{Method}} &
            \multicolumn{2}{c}{\textbf{\pc{50}}} &
            \multicolumn{2}{c}{\textbf{\pc{60}}} &
            \multicolumn{2}{c}{\textbf{\pc{70}}} &
            \multicolumn{2}{c}{\textbf{\pc{80}}} &
            \multicolumn{2}{c}{\textbf{\pc{90}}} \\
            \cmidrule(lr){3-4}
            \cmidrule(lr){5-6}
            \cmidrule(lr){7-8}
            \cmidrule(lr){9-10}
            \cmidrule(lr){11-12}
            & & Wiki & C4 & Wiki & C4 & Wiki & C4 & Wiki & C4 & Wiki & C4 \\
            \midrule

            \multirow{6}{*}{\makecell[c]{OPT-125M\\
            {\scriptsize Dense: 27.60 / 26.53}\\
            {\scriptsize 5M delta budget}}}
            & \magnitude
            & 200.3 & 146.9 & 937.9 & 609.3 & 3855 & 2323 & 4882 & 3209 & 6513 & 4355 \\
            & \wanda
            & 38.99 & 34.85 & 75.51 & 62.19 & 344.4 & 253.8 & 1873 & 1007 & 4333 & 3113 \\
            & \sparsegpt
            & 36.98 & 33.55 & 58.76 & 48.92 & 233.6 & 161.3 & 2090 & 1142 & 6987 & 3418 \\
            & \alps
            & 32.73 & 30.92 & 43.13 & 37.13 & 89.29 & 65.68 & 493.65 & 265.35 & 2720 & 1142 \\
            & \elsa
            & 36.44 & 31.55 & 39.42 & 33.32 & 47.56 & 38.83 & 62.63 & 46.42 & 95.92 & 61.68 \\
            & \textbf{\etaps}
            & \textbf{30.79} & \textbf{28.17}
            & \textbf{35.14} & \textbf{30.62}
            & \textbf{42.85} & \textbf{35.06}
            & \textbf{58.96} & \textbf{42.61}
            & \textbf{79.84} & \textbf{55.44} \\
            \midrule

            \multirow{6}{*}{\makecell[c]{OPT-1.3B\\
            {\scriptsize Dense: 14.61 / 16.06}\\
            {\scriptsize 28M delta budget}}}
            & \magnitude
            & 1818 & 412.8 & 8692 & 4766 & 9592 & 6662
            & 1.6e4 & 1.1e4 & 2.7e4 & 1.7e4 \\
            & \wanda
            & 18.39 & 20.62 & 26.54 & 28.34 & 101.0 & 92.53
            & 2805 & 1423 & 1.3e4 & 7382 \\
            & \sparsegpt
            & 17.43 & 19.22 & 22.06 & 23.13 & 49.91 & 45.31
            & 954.6 & 363.9 & 6416 & 2783 \\
            & \alps
            & 16.86 & 18.57 & 20.40 & 21.49 & 35.98 & 34.09
            & 260.1 & 161.6 & 4502 & 1821 \\
            & \elsa
            & 19.55 & 18.60 & 21.94 & 19.91 & 25.32 & 23.32
            & 36.56 & 29.50 & 60.96 & 42.99 \\
            & \textbf{\etaps}
            & \textbf{14.84} & \textbf{16.65}
            & \textbf{16.70} & \textbf{18.00}
            & \textbf{20.26} & \textbf{20.33}
            & \textbf{29.40} & \textbf{26.61}
            & \textbf{54.64} & \textbf{42.30} \\
            \midrule

            \multirow{6}{*}{\makecell[c]{Gemma-2-2B\\
            {\scriptsize Dense: 8.71 / 13.16}\\
            {\scriptsize 42M delta budget}}}
            & \magnitude
            & 54.04 & 59.54
            & 2231 & 1919
            & 2.6e7 & 2.8e6
            & 1.8e9 & 3.1e8
            & 2.2e10 & 1.8e10 \\
            & \wanda
            & 13.44 & 19.52
            & 33.63 & 47.73
            & 212.4 & 227.3
            & 3003 & 1666
            & 1.8e4 & 1.6e4 \\
            & \sparsegpt
            & 13.01 & 18.57 & 21.25 & 29.39 & 65.97 & 80.75
            & 524.4 & 439.9 & 6907 & 3129 \\
            & \alps
            & 10.94 & 15.75 & 14.39 & 20.27 & 25.25 & 35.49
            & 74.87 & 94.07 & 255.0 & 248.1 \\
            & \elsa
            & 12.52 & 16.88 & 16.53 & 21.71 & 20.51 & 23.79
            & 28.70 & 30.60 & 41.94 & 40.87 \\
            & \textbf{\etaps}
            & \textbf{10.33} & \textbf{14.08}
            & \textbf{12.49} & \textbf{16.14}
            & \textbf{15.67} & \textbf{18.74}
            & \textbf{21.88} & \textbf{24.23}
            & \textbf{34.74} & \textbf{34.84} \\
            \midrule

            \multirow{6}{*}{\makecell[c]{LLaMA-3.2-3B\\
            {\scriptsize Dense: 7.81 / 11.33}\\
            {\scriptsize 49M delta budget}}}
            & \magnitude
            & 138.6 & 214.8
            & 1.5e4 & 1.4e4
            & 1.0e5 & 8.1e4
            & 3.5e5 & 3.5e5
            & 3.0e5 & 2.3e5 \\
            & \wanda
            & 12.99 & 19.04
            & 31.23 & 42.57
            & 146.1 & 183.5
            & 2287 & 1148
            & 1.3e4 & 8673 \\
            & \sparsegpt
            & 12.26 & 17.39 & 22.61 & 29.96 & 68.92 & 80.36
            & 268.7 & 226.0 & 1498 & 1132 \\
            & \alps
            & 11.38 & 15.95 & 18.26 & 23.00 & 41.20 & 44.99
            & 128.7 & 105.6 & 538.4 & 369.1 \\
            & \elsa
            & 12.16 & 17.05 & 15.82 & 20.68 & 23.27 & 26.65
            & 34.12 & 35.78 & 49.58 & 48.39 \\
            & \textbf{\etaps}
            & \textbf{10.14} & \textbf{13.97}
            & \textbf{12.85} & \textbf{16.61}
            & \textbf{17.12} & \textbf{19.93}
            & \textbf{23.82} & \textbf{25.48}
            & \textbf{36.01} & \textbf{35.15} \\
            \midrule

            \multirow{6}{*}{\makecell[c]{LLaMA-2-7B\\
            {\scriptsize Dense: 5.47 / 7.26}\\
            {\scriptsize 80M delta budget}}}
            & \magnitude
            & 16.00 & 21.31
            & 1920 & 2071
            & 5.0e4 & 2.8e4
            & 1.1e5 & 1.1e5
            & 4.8e4 & 5.4e4 \\
            & \wanda
            & 6.92 & 9.24 & 10.79 & 14.02 & 77.08 & 81.51
            & 2226 & 2010 & 6477 & 5099 \\
            & \sparsegpt
            & 7.00 & 9.23 & 10.20 & 12.93 & 27.77 & 31.37
            & 119.4 & 100.7 & 1369 & 1033 \\
            & \alps
            & 6.85 & 9.01 & 9.28 & 11.33 & 18.88 & 20.18
            & 48.20 & 47.21 & 232.2 & 174.5 \\
            & \elsa
            & 7.27 & 9.44 & 8.36 & 10.61 & 12.42 & 13.69
            & 17.99 & 17.34 & 27.98 & 23.20 \\
            & \textbf{\etaps}
            & \textbf{6.66} & \textbf{8.20}
            & \textbf{7.69} & \textbf{9.04}
            & \textbf{10.99} & \textbf{10.87}
            & \textbf{14.17} & \textbf{13.09}
            & \textbf{20.98} & \textbf{17.39} \\
            \midrule

            \multirow{6}{*}{\makecell[c]{LLaMA-2-13B\\
            {\scriptsize Dense: 4.88 / 6.72}\\
            {\scriptsize 125M delta budget}}}
            & \magnitude
            & 6.82 & 9.37
            & 11.82 & 14.64
            & 218.1 & 196.4
            & 3.9e4 & 4.8e4
            & 7.4e4 & 6.4e4 \\
            & \wanda
            & 5.97 & 8.29
            & 8.39 & 11.53
            & 45.19 & 55.31
            & 1127 & 877.2
            & 2.0e4 & 1.1e4 \\
            & \sparsegpt
            & 6.02 & 8.22
            & 8.25 & 10.96
            & 19.65 & 23.25
            & 94.64 & 78.75
            & 1294 & 911.9 \\
            & \alps
            & \textbf{5.92} & 8.10
            & 7.54 & 9.91
            & 14.13 & 16.35
            & 38.48 & 36.99
            & 241.2 & 149.8 \\
            & \elsa
            & 6.54 & 8.78
            & 7.96 & 9.93
            & 11.14 & 12.20
            & 17.21 & 16.60
            & 30.19 & 25.07 \\
            & \etaps
            & \textbf{5.92} & \textbf{7.43}
            & \textbf{6.95} & \textbf{8.13}
            & \textbf{9.37} & \textbf{9.29}
            & \textbf{11.93} & \textbf{11.10}
            & \textbf{17.50} & \textbf{14.89} \\

            \bottomrule
        \end{tabular}%
    }
\end{table}

\end{document}